%% file: main.tex
\documentclass[11pt,a4paper]{article}
\usepackage[inner=2.5cm,outer=2.5cm,top=2.6cm,bottom=2.7cm,headheight=24pt]{geometry}
\usepackage{fontspec}
\usepackage{amsmath,amssymb,booktabs,longtable,array,tabularx,multirow}
\usepackage{graphicx,xcolor,enumitem,placeins,float}
\usepackage[numbers,sort&compress]{natbib}
\usepackage[colorlinks,breaklinks]{hyperref}
\usepackage{xurl}
\input{latex/style}
\input{latex/authors}
\graphicspath{{figures/}}
\setlist{nosep,leftmargin=2em,topsep=0.3em}
\newcolumntype{Y}{>{\raggedright\arraybackslash}X}
\newcolumntype{P}[1]{>{\raggedright\arraybackslash}p{#1}}
\renewcommand{\arraystretch}{1.18}
\newcommand{\chapterstart}[1]{\par\FloatBarrier\Needspace{9\baselineskip}\section{#1}}
\newcommand{\Frontier}{\textit{Frontier}}
\newcommand{\Guided}{\textit{Principle-guided}}
\newcommand{\KL}{\operatorname{KL}}
\newcommand{\CE}{\operatorname{CE}}
\title{Data-Efficient Language Modeling: From Frontier\\
Advancement to Principle-Guided Model Improvement}
\newcommand{\ReportSubtitle}{Qiushi Engine's End-to-End Autonomous Research on BabyLM 2026 Strict-Small}
\date{8 September 2026}
\hypersetup{pdftitle={Data-Efficient Language Modeling: From Frontier Advancement to Principle-Guided Model Improvement},pdfauthor={Qiushi Engine Team},pdfsubject={End-to-end autonomous research on BabyLM 2026 Strict-Small}}
\begin{document}
\maketitle
\thispagestyle{plain}
\begin{abstract}
Learning from limited text requires more than repeated exposure: models must learn to use relevant information, apply what they learn to new inputs, and retain useful capabilities during further training. Qiushi Engine conducted a long-horizon, end-to-end autonomous research program on BabyLM 2026 Strict-Small, under limits of ten million corpus words and one hundred million cumulative word presentations. Three complete research stages connected frontier advancement, principle discovery, and principle-guided model improvement, with autonomous literature research, method development, experimentation, analysis, and synthesis throughout. Stage I combined compact restatements, budget reinvestment, and residual incremental learning to build a frontier model. Stage II investigated experience organization, supervision, capability reuse and retention, learning dynamics, and measurement. Exact repetition and aligned restatement produced different patterns of context use, depending on the target relation and prediction window. In controlled tasks, recovering familiar performance did not ensure that unseen inputs could still use a learned computation. These findings support a testable data-efficient learning principle: organize experience around the contextual information and relationships required for prediction; separately design visible information, supervised targets, and functional preservation; and test whether the intended capability is learned, generalizes to new inputs, and survives further training. Stage III applied these findings by retaining source text, masking more local clues, supervising selected targets, and preserving predictions on ordinarily masked inputs. Both continuations from the same parent outperformed ordinary continued training on the complete nine-metric evaluation. Overall advanced from 42.02 to 42.25 across the two model generations, with the second achieving the highest score in the public Strict-Small snapshot of 8 September 2026. Independent studies yielded further findings and methods on compression, relational anchors, shared representations, and measurement controls. Models are available on Hugging Face; code, data construction, evaluations, and research records accompany the GitHub repository. Together, these stages provide a concrete instance of Research RSI---recursive self-improvement of the research process. Scientific understanding, method innovations, and experimental experience from earlier research change subsequent questions and designs; new experiments test and refine them. A frontier advance thus becomes both a better model and a basis for improving how further research is pursued.
\end{abstract}
{\setstretch{1.05}\tableofcontents\par}
\clearpage
\input{chapters/01_background}
\input{chapters/03_frontier}
\input{chapters/04_principles}
\input{chapters/05_guided_improvement}
\input{chapters/06_research_lineage}
\input{chapters/07_discussion}
\input{chapters/08_conclusion}
\par\FloatBarrier\clearpage
\appendix
\input{chapters/appendices}
\input{chapters/research_catalog}

\par\FloatBarrier\clearpage
\begingroup\small\setstretch{1.05}\setlength{\bibsep}{2pt}
\bibliographystyle{unsrtnat}
\bibliography{references}
\endgroup
\end{document}

%% file: latex/style.tex
\usepackage{setspace,etoolbox,needspace,caption,fancyhdr,titlesec}
\renewenvironment{abstract}{\par\begingroup\small\setstretch{1.04}\noindent\textbf{Abstract}\par\smallskip\noindent}{\par\endgroup\medskip}
\definecolor{ReportPrimary}{HTML}{1F3A5F}
\definecolor{ReportText}{HTML}{202328}
\definecolor{ReportRule}{HTML}{D6DDE5}
\definecolor{QiushiBlue}{HTML}{003F88}
\newlength{\ReportFigureWidth}
\newcommand{\ReportTableContinuation}{\textbf{\tablename~\thetable}\quad (continued)}
\pretocmd{\tableofcontents}{\clearpage}{}{}
\renewcommand{\footrulewidth}{0.4pt}
\renewcommand{\footrule}{\hbox to\headwidth{\color{ReportRule}\leaders\hrule height \footrulewidth\hfill}}
\fancypagestyle{plain}{\fancyhf{}\fancyfoot[L]{\small\color{QiushiBlue}\textbf{Qiushi Engine}}\fancyfoot[R]{\small\thepage}\renewcommand{\footrulewidth}{0.4pt}}
\titleformat{\section}{\Large\bfseries\color{ReportText}}{\thesection}{0.65em}{}
\titleformat{\subsection}{\large\bfseries\color{ReportText}}{\thesubsection}{0.65em}{}
\titleformat{\subsubsection}{\normalsize\bfseries\color{ReportText}}{\thesubsubsection}{0.65em}{}
\titlespacing*{\section}{0pt}{16pt plus 2pt minus 2pt}{7pt}
\titlespacing*{\subsection}{0pt}{11pt plus 2pt minus 1pt}{5pt}
\titlespacing*{\subsubsection}{0pt}{8pt plus 1pt}{3pt}
\AtBeginEnvironment{longtable}{\small\setstretch{1.04}\renewcommand{\arraystretch}{1.10}}
\hypersetup{linkcolor=ReportText,citecolor=ReportText,urlcolor=ReportPrimary}
\makeatletter
\renewcommand{\@maketitle}{%
  \begingroup\setstretch{1.04}\parindent=0pt
  \noindent\begin{minipage}[c]{0.64\linewidth}
    \includegraphics[width=42mm]{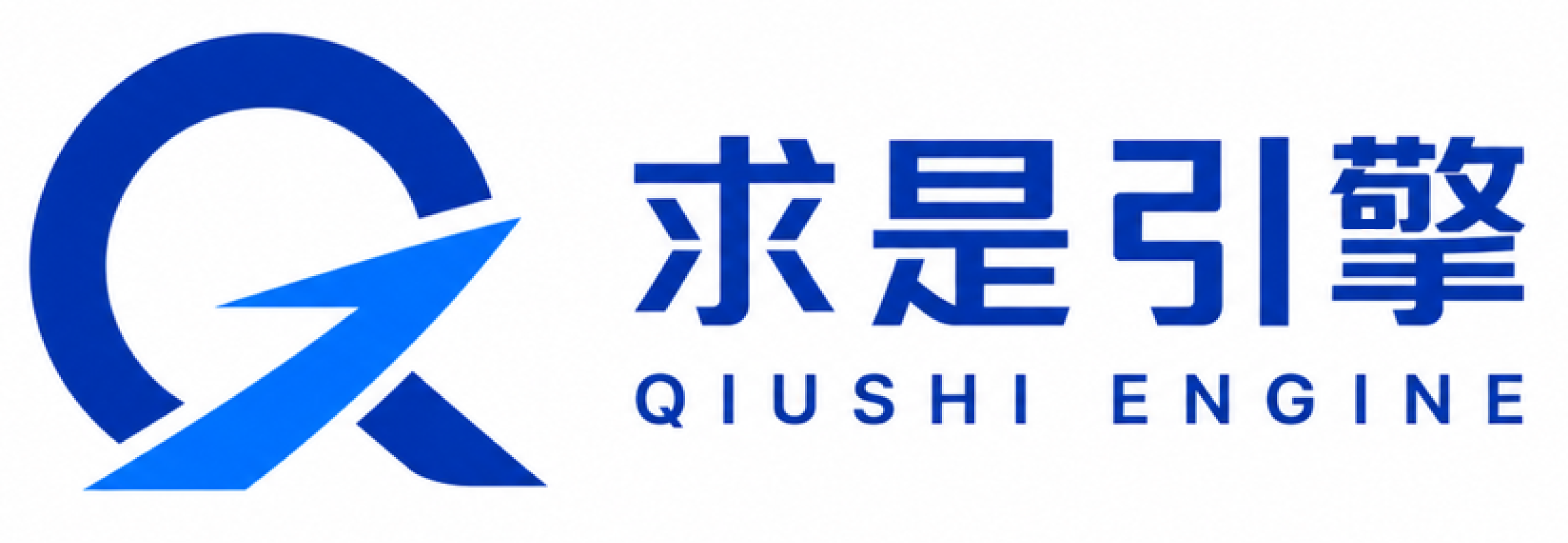}%
  \end{minipage}\hfill
    \begin{minipage}[c]{0.32\linewidth}\raggedleft\small\color{ReportText}\@date\end{minipage}\par
  \vspace{3mm}{\color{ReportRule}\rule{\linewidth}{0.5pt}}\par
  \vspace{4mm}{\centering\fontsize{16}{20}\selectfont\bfseries\@title\par}
  \vspace{2.5mm}{\centering\small\ReportSubtitle\par}
  \vspace{3mm}{\centering\small\ReportAuthorBlock\par}
  \vspace{1.5mm}{\centering\small\ReportAffiliation\par}
  \vspace{1mm}{\centering\small\mbox{\ReportCorrespondence}\par}
  \vspace{2mm}\endgroup
}
\makeatother

%% file: latex/authors.tex
\newcommand{\ReportAuthorBlock}{%
\renewcommand{\arraystretch}{1.16}%
\begin{tabular*}{0.98\linewidth}{@{\extracolsep{\fill}}cccc@{}}
Shuxing Yang & Kaihao Zhu & Junjie Yang & Rui Zhao\\
Junyao Wu & Yize Wang & Wenhao Li & Fujia Chen\\
Taowen Deng & Shenzhan Hong & Yaqi Li & Zichen Li\\
Jincheng Mi & Yuang Pan & Hongsheng Chen\textsuperscript{*} & Yihao Yang\textsuperscript{*}
\end{tabular*}}
\newcommand{\ReportAffiliation}{College of Information Science and Electronic Engineering, Zhejiang University\\Qiushi Engine Team}
\newcommand{\ReportCorrespondence}{\textsuperscript{*}Correspondence: Yihao Yang (\href{mailto:yangyihao@zju.edu.cn}{yangyihao@zju.edu.cn}); Hongsheng Chen (\href{mailto:hansomchen@zju.edu.cn}{hansomchen@zju.edu.cn})}

%% file: chapters/01_background.tex
\chapterstart{BabyLM 2026: Learning from Limited Data}
\label{sec:intro}

Large-scale training has shown how far language-model capabilities can grow with additional resources. Learning from limited text asks a complementary question: how much more can a model learn from the experience already available to it? A better method must do more than fit that text. It should help the model use relevant information, apply what it learns to new expressions, and retain useful capabilities as training continues.

BabyLM provides shared experimental conditions for this question. This report concerns its Strict-Small track and the complete research program conducted by Qiushi Engine: building a frontier model, investigating the conditions behind its behavior, and using those findings to improve the next model. Model scores are one outcome. The resulting methods and testable explanations are another.

\subsection{The Scientific Problem of Learning from Limited Data}

Language models learn by predicting text. Training compares a model's predicted probabilities with the words that actually occur, measures the error through a loss, and adjusts the parameters. In this report, \emph{supervision} specifies which predictions contribute to learning: a target is a token to be recovered, and a target position identifies where that token occurs. Repeated training can establish lexical, syntactic, semantic, and contextual regularities. The scientific question is which regularities actually support a correct answer.

Suppose a passage states that a key was moved from a drawer to a box, and a later query asks where the key is now. The required computation uses the change of state. A model that instead relies on the familiar association between ``key'' and ``drawer'' may fit many ordinary sentences without learning that operation. If nearby words usually reveal the target, training offers little incentive to consult earlier information. \textbf{Encountering relevant information and learning to use it are distinct achievements.}

This distinction matters especially when data are scarce. Adding a new passage, repeating an old one, and restating its content in different words all consume training budget, but need not provide interchangeable learning opportunities. Repetition can consolidate a pattern; restatement can vary its expression; a new source can broaden coverage. Their value depends on the model's current capabilities and remaining training budget. Data-constrained scaling studies likewise examine repeated exposure jointly with the amount of nonrepeated data and the training configuration~\citep{muennighoff2025scaling}.

Data-efficient language modeling studies how to use limited learning material more effectively. Parameter count, corpus size, cumulative exposure, and compute measure different resources: capacity, available text, repeated presentations, and execution cost. A small model can read a large amount of text repeatedly; a larger model can have limited experience. BabyLM's strict English tracks constrain both the corpus and cumulative exposure, allowing methods to be compared under common data limits.

Our central question is: \textbf{what training experience teaches a model to use information relationships in context, and what allows that ability to remain usable during further learning?} We use \emph{contextual dependencies} to mean the contextual information and relationships required to make a prediction. This question connects language acquisition, compositional generalization---using learned elements in new combinations---and continual learning---acquiring new capabilities while retaining previous ones. Natural text, controlled relational tasks, and full language-model training examine different parts of this question.

\subsection{Research Background and Track Rules}
\label{sec:setting}

Launched in 2023, BabyLM enters its fourth edition in 2026 as a shared task and workshop associated with EMNLP 2026. It brings together natural language processing, computational cognitive science, and language-acquisition research through training data, evaluation software, baseline models, and public comparisons~\citep{warstadt2023babylm,babylm2026website}. Earlier editions have established a body of work on data processing, model design, and learning behavior~\citep{charpentier2025babylm}.

Its value lies in the experimental question as much as the leaderboard. For machine learning, BabyLM provides a tractable setting for comparing training methods. For language and cognition, it provides models whose input and learning trajectories can be manipulated. For research teams with limited resources, it makes training from scratch and retaining intermediate models more feasible. The task does not require a complete simulation of a child; it makes learning from restricted input a shared object of investigation.

The 2026 tracks include English Strict-Small, English Strict, and Multilingual, covering English, Dutch, and Chinese. Results in this report belong only to Strict-Small. The two English budget levels are summarized in Table~\ref{tab:tracks}~\citep{choshen2026babylm}.

\begin{table}[htbp]
\caption{Data limits for the English tracks. Word counts follow the track rules, rather than the model tokenizer. M denotes million and B denotes billion.}
\label{tab:tracks}
\begin{tabularx}{\linewidth}{@{}lYY@{}}
\toprule Track & Corpus limit & Cumulative exposure limit\\\midrule
Strict-Small (this study) & 10M words & 100M words\\
Strict & 100M words & 1B words\\
\bottomrule
\end{tabularx}
\end{table}

The corpus is the collection available for training; cumulative exposure counts the text actually presented during training. Reading a thousand-word article ten times still uses one article but consumes ten thousand words of exposure. Rewritten text, repeated passages, and additional inputs used to preserve existing behavior must also be accounted for under the applicable rules. A tokenizer converts text into computational units called tokens. Because one word can produce several tokens, tokenizer counts cannot replace the track's word counts.

Intermediate checkpoints---model weights saved at successive training amounts---are required for evaluating learning trajectories. External teachers and auxiliary training methods have additional rules; satisfying the final word count alone is insufficient to document them. Appendix~\ref{app:repro} records the data, teacher roles, and training settings used here. The evaluation therefore concerns both the final score and the experience through which it was obtained.

A fixed budget still leaves substantial scientific freedom. Data content, expression, vocabulary, architecture, objectives, learning rate, batch size, presentation order, and model selection all remain consequential choices. The learning rate controls the magnitude of parameter updates; the batch size controls how much evidence is combined in an update. At a fixed exposure budget, larger batches can also mean fewer updates. Strict-Small supplies common constraints, not a prescribed training recipe.

\subsection{Evaluating Language Abilities and Learning Behavior}

No single test establishes grammatical knowledge, state tracking, commonsense use, transfer to new tasks, and correspondence with human language behavior. BabyLM's nine top-level metrics measure these separately. \textbf{The overall score combines task performance with measures of human behavioral correspondence.} Each component must be interpreted through its own task, scale, and reference level.

\begin{table}[htbp]
\caption{The nine top-level English evaluation metrics. The first seven form NLP Average; the final two form Human-like Average. Scores use the fixed official-compatible evaluation implementation.}
\label{tab:metrics}
\small
\begin{tabularx}{\linewidth}{@{}P{29mm}Y Y@{}}
\toprule Metric & Main question & Interpretation\\\midrule
BLiMP & Does the model prefer the grammatical member of a minimal pair? & Choice accuracy across grammatical phenomena.\\
BLiMP Supplement & Does it handle additional linguistic phenomena, including question answering and reference? & Aggregation of supplementary task groups.\\
EWoK & Does contextual information support elementary world knowledge? & Relative plausibility of continuations under different contexts.\\
Entity Tracking & Can it recover an object's state after moves or updates? & Candidate-answer accuracy after state changes.\\
COMPS & Can properties transfer to new concepts despite distractors? & Property, inheritance, and interference tests.\\
GlobalPIQA & Can it judge everyday physical actions and uses? & Choice accuracy across subsets with different formats.\\
(Super)GLUE & Do pretrained representations support downstream understanding? & Mean task metrics after prescribed fine-tuning.\\
Reading & Do model predictions explain variation in human reading? & Explanatory gain beyond word length, frequency, and other controls.\\
AoA & Does the learning order of words correspond to children's acquisition? & A trajectory-based correlation measure.\\
\bottomrule
\end{tabularx}
\end{table}

\subsubsection{Linguistic Form, Knowledge, and Context Use}

BLiMP isolates small grammatical contrasts such as subject--verb agreement. EWoK changes the context and asks whether the model's judgment changes accordingly. COMPS tests whether a new concept inherits the appropriate property and whether unrelated concepts interfere. Entity Tracking requires the model to recover a state after a sequence of operations~\citep{warstadt2020blimp,ivanova2025ewok,misra2023comps,kim2023entity}. These tasks separate plausible language from the use of the information needed for a particular answer.

``The keys are on the table'' illustrates agreement; a key moved from a drawer into a box illustrates state updating; assigning a newly named animal to a category illustrates property inheritance. These are explanatory examples, not benchmark items. They show why the challenge extends beyond remembering facts: limited experience must support different structural and relational operations.

Most of these tests score sentences or candidate answers with the pretrained model, without training on the individual questions. This is zero-shot evaluation. Chance performance depends on the task format. The GlobalPIQA English evaluation used here has 103 four-choice items and 100 two-choice items, with equal weight assigned to the two subsets; uniform random choice therefore has an expected score of 37.5. A score near 40 cannot be interpreted without that reference~\citep{chang2026globalpiqa,babylm2026evaluation}.

\subsubsection{Transfer Through Downstream Fine-Tuning}

(Super)GLUE asks whether pretrained representations support language understanding after task-specific training, or fine-tuning. BabyLM uses seven selected tasks covering yes/no questions, entailment, paraphrase, multi-sentence reading judgments, and reference resolution~\citep{wang2018glue,wang2019superglue}. It tests whether pretraining supplies a useful basis for subsequent learning.

Some tasks use accuracy and others F1. Accuracy measures the proportion of correct answers; F1 combines precision and recall for the relevant predictions. The score is consequently neither the full original GLUE/SuperGLUE suite nor a pooled accuracy over all questions. Fine-tuning seeds, classification heads, training settings, and model selection must be kept consistent when comparing models.

\subsubsection{Reading Behavior and Word Acquisition}

Reading measures how model expectations relate to human reading, rather than accuracy on reading-comprehension questions. A less probable word has higher \emph{surprisal}, defined as its negative log probability given the preceding context. The evaluation asks whether surprisal explains eye movements or word-by-word reading times beyond factors such as length and frequency~\citep{devarda2024cloze}.

The score follows the BabyLM 2026 evaluation implementation~\citep{babylm2026evaluation}. If control variables explain a proportion $R_0^2$ of the variance and adding model predictions raises this to $R_1^2$, the normalized gain is
\begin{equation}
q=\frac{R_1^2-R_0^2}{1-R_0^2}.
\end{equation}
An increase from 20\% to 24\% explained variance gives $0.04/0.80=0.05$, or 5 on a percentage scale. Small Reading values can therefore represent meaningful explanatory gains. The leaderboard aggregates the prescribed reading responses~\citep{babylm2026leaderboard}.

AoA denotes Age of Acquisition. The evaluation estimates word-learning progress from prediction curves across checkpoints and compares it with children's acquisition order~\citep{chang2022word}. The implementation used here computes Pearson correlation, returning zero when fewer than three words have valid estimates or the correlation test gives $p>0.1$; a negative correlation that passes this threshold remains negative~\citep{babylm2026evaluation}. Appendix~\ref{app:evaluation} specifies the scoring version and checkpoint trajectories. \textbf{An AoA score of zero does not mean that the model learned no words.} Reading and AoA make behavioral correspondence an explicit dimension alongside task competence.

\subsection{Aggregate Scores and the Public Leaderboard}

Let $m_1,\ldots,m_9$ be the component scores, $N$ the mean of the seven language-task scores, and $H$ the mean of the two human-behavior scores. Overall Average is
\begin{equation}
O=\frac{1}{9}\sum_{j=1}^{9}m_j=\frac{7N+2H}{9}.
\label{eq:overall}
\end{equation}
Each component has equal weight; NLP and Human-like therefore have unequal total weights. Overall permits comparison under a shared protocol. It is not the percentage of language a model understands, and it cannot be compared directly with large-model scores from different tests and training budgets.

The two representative models are \nolinkurl{Qiushi-Engine-Frontier-Advancement} and \nolinkurl{Qiushi-Engine-Principle-Guided-Frontier-Advancement}, abbreviated \Frontier{} and \Guided{}. Their public Overall scores are 42.02 and 42.25. The latter is the highest Overall in the retained Strict-Small snapshot.

Table~\ref{tab:board} includes these two models and the eight highest-Overall entries from other publishers, without the team's historical submissions or an adjusted ranking column. The snapshot was recorded on 8 September 2026 at 20:53 Beijing time. This is a comparison of public submissions, distinct from the determination of workshop awards~\citep{babylm2026leaderboard}.

\begin{table}[htbp]
\caption{Two representative models and eight external public submissions. Readable names are used here; full identities and links accompany the data. Scores retain the leaderboard's displayed precision.}
\label{tab:board}
\small
\begin{tabularx}{\linewidth}{@{}Yrrr@{}}
\toprule Publisher/model & Overall & NLP & Human-like\\\midrule
\input{tables/leaderboard}
\end{tabularx}
\end{table}

\begin{figure}[htbp]
\centering\makebox[\linewidth][c]{\includegraphics[width=\ReportFigureWidth]{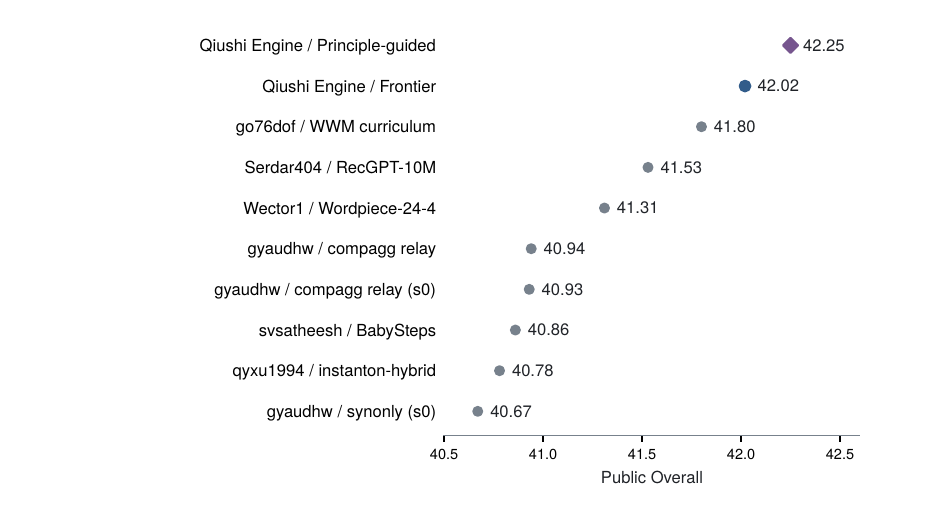}}
\caption{Public Overall scores on BabyLM 2026 Strict-Small, 8 September 2026. Points and labels show the displayed scores for the two Qiushi Engine models and the eight external submissions in Table~\ref{tab:board}.}
\label{fig:leaderboard}
\end{figure}

The external entries lie between 40.67 and 41.80, a span of 1.13 points, with adjacent differences as small as 0.01. The second-generation model exceeds the highest external displayed score by 0.45. Closely spaced totals make component-level analysis important: which abilities improved, which incurred costs, and whether repeated training from the same starting model retains the advantage. The leaderboard establishes the public comparison; common-parent experiments below test the training method itself.

Similar totals also conceal different capability profiles. The highest NLP score in the comparison is 53.59, yet weaker human-behavior scores prevent that model from achieving the highest Overall. BLiMP ranges from 67.13 to 73.11, Entity Tracking from 16.59 to 29.40, and GlobalPIQA from 36.08 to 40.68, near its 37.5 chance reference. Syntactic preferences, changing states, and physical commonsense remain unevenly learned. Appendix~\ref{app:board} retains all nine components.

\subsection{The Main Challenges}

\textbf{Coverage versus practice.} A fixed word budget can buy broader content or more rehearsal. Compression may create room for additional examples while removing relational clues. Rewriting may provide useful variation or redundancy. Data quality must be judged by what the transformation teaches, not text volume or fluency alone.

\textbf{Correct prediction versus relevant information use.} A model may complete a target using nearby collocations without reading the source. Loss reduction can be genuine even when the desired relational ability has not formed. Keeping targets fixed while changing visible clues, or keeping text fixed while changing supervised positions, separates these possibilities.

\textbf{New learning versus retention.} Training a particular relation can improve that task while degrading ordinary language abilities. Familiar-item performance may recover even when new names cannot use the learned computation. An integrated improvement requires measuring acquisition and retention together.

\textbf{Interactions among representation, architecture, and optimization.} Tokenization determines the input units; architecture determines the available computations; objectives and optimization determine which relationships are practiced. Early rankings can reverse, and the same random seed does not ensure matching initial parameters across architectures. Useful controls isolate the differences relevant to the proposed explanation.

\textbf{Training a model versus conducting the research.} The cost of one final training run is not the cost of finding its method. Candidate selection, mechanism experiments, full evaluation, and error analysis are additional work. Autonomous research must connect these activities over time: preserve reliable starting points, identify informative questions, implement controls, correct mistakes, revise explanations, and construct another executable method. This complete process is the relevant setting for assessing Qiushi Engine's research contribution.

\subsection{Performance Gains and Scientific Contributions}

Because Overall averages nine metrics, a 0.9-point increase in one component with no other change yields a 0.1-point increase in Overall. Several smaller changes can yield the same total. A narrow aggregate margin may therefore reflect identifiable behavioral changes, but sample size, training randomness, and model selection also matter. Component results and matched comparisons establish what the margin means.

The study compares the new methods with ordinary continuation from the same first-generation model. Ordinary continuation achieves 42.0926 and 42.1159 across two continuation seeds. Dense masking with sparse supervision reaches 42.2025 and 42.1789 at the same cumulative word exposure. Adding ordinary-input preservation yields 42.2464 and 42.2317, with extra preservation presentations accounted for separately.

The scientific contribution extends beyond these differences. Stage I produced a way to compress expression and reinvest the word budget, an extensible model structure, and a complete training and evaluation implementation. Stage II established relation-dependent context-use effects and separated familiar performance from reuse on unseen inputs. Stage III converted those findings into training operations and tested their practical value. \textbf{New observations, discriminating explanations, executable methods, and evidence together define the contribution.}

Independent investigations studied how sparse labels identify relational structure, how new names become associated with known objects, and how explicit memory retrieves the appropriate state. These yielded work on relational anchors, identity matching, and entity memory, developed in Section~\ref{sec:lineage}.

\subsection{Research Value and Prospective Applications}

Data-limited research produces methods that others can inspect and adapt. Specialized domains and low-resource languages often lack sufficient high-quality training text. Preserving relevant content, organizing corresponding expressions, selecting supervision, and continuing learning are therefore plausible transfer directions. The present experiments provide concrete methods and hypotheses for those applications, rather than an already established cross-domain result.

Small-scale experiments also make mechanisms easier to study. Data and training variables can be controlled directly, and changes to a method can be tested economically. Local prediction shortcuts, supervision imbalance, and excessive preservation constraints suggest specific questions for larger models: when do the same conditions recur, and how do they vary with scale, task, and budget?

For language and cognitive science, the model is an experimental object whose experience can be manipulated and whose learning can be followed. Joint evaluation of grammar, conceptual inheritance, reading behavior, and acquisition order helps identify when better task performance coincides with more human-like behavior and when the two diverge.

For autonomous AI research, the contribution is a sustained scientific process. Qiushi Engine turns differences in trained-model behavior into questions, constructs comparisons, revises explanations, and uses the resulting understanding in subsequent model design. Its models, methods, and scientific judgments develop together. We describe the reuse of this accumulation in subsequent research as Research RSI---recursive self-improvement of the research process. Section~\ref{sec:rsi} develops its concrete realization, relationship to prior work, and empirical scope.

\subsection{Three Complete Research Stages Conducted by Qiushi Engine}

Qiushi Engine conducted the research across all three stages: data preparation, representation and architecture exploration, objective design, experimental implementation, training, evaluation, mechanism analysis, and synthesis of the findings. The team supplied goals, resources, and stage-level requirements; Qiushi Engine carried out the scientific work within that setting. Computational research used the AI Lab skill and associated scientific programs on two NVIDIA H100 GPUs and CPUs. Section~\ref{sec:research_organization} describes the research organization; Appendix~\ref{app:compute} records the compute environment.

Each stage is a complete research process, including questions, hypotheses, method construction, experiments, analysis, and reporting. The stages differ in their scientific aims, while data, models, experimental methods, and findings connect them.

\textbf{Stage I: frontier advancement.} Qiushi Engine explored text construction, tokenization, representation, architecture, optimization, and training schedules under limited data. Compact restatements, budget reinvestment, and residual incremental learning became a combined method that produced the first frontier model. Its training also exposed changing method rankings and differences between local learning and broad performance.

\textbf{Stage II: principle discovery.} Qiushi Engine treated those models and observations as research objects. It proposed competing explanations and tested them through interventions on data, windows, supervision, representations, and internal computations. The work covered experience structure and compression, inductive biases, learning signals, retention, curricula, learning dynamics, and measurement. Relation learning and capability reuse informed the next model; anchors, identity matching, initialization controls, and measurement methods also produced independent findings.

\begin{figure}[H]
\centering\makebox[\linewidth][c]{\includegraphics[width=\ReportFigureWidth]{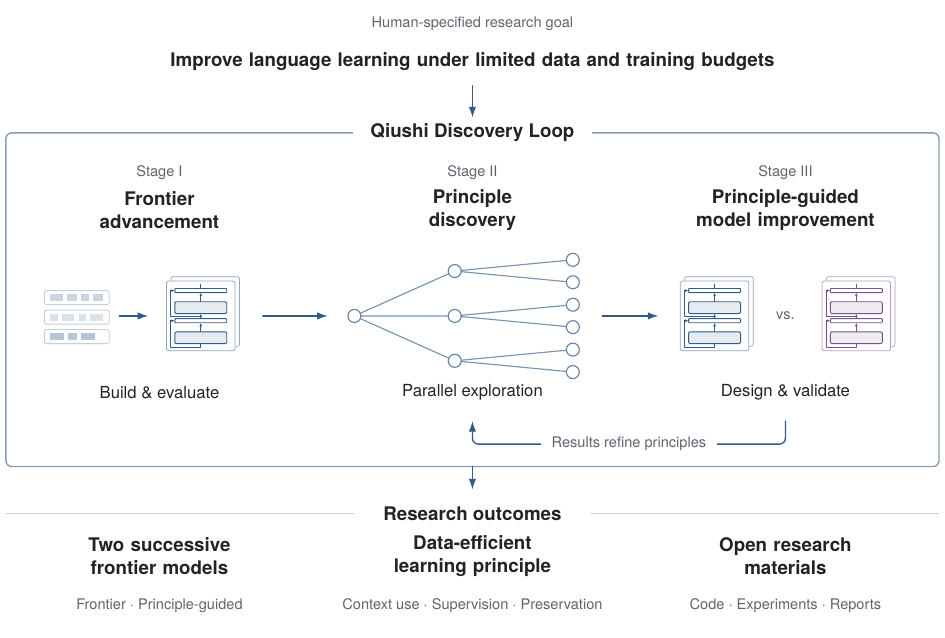}}
\caption{Qiushi Engine's three-stage autonomous research program. Frontier advancement, principle discovery, and principle-guided model improvement each comprise a complete research process. Models, methods, and scientific understanding develop across the stages, producing two successive frontier models, data-efficient learning principles, and open research materials.}
\label{fig:program}
\end{figure}

\textbf{Stage III: principle-guided model improvement.} Using the first model and its existing text, Qiushi Engine changed the visible input, supervised positions, and preservation constraints. Ordinary continuation, target-set comparisons, mechanism tests, and full evaluation established the second model and refined earlier explanations of supervision quantity and preservation strength. Figure~\ref{fig:program} summarizes the continuity of the three research processes.

The resulting design principle is: \textbf{organize limited experience around the contextual dependencies a capability requires; make relevant information available during prediction, train the model to use it, and evaluate acquisition, reuse, and retention together.} Acquisition asks whether the target behavior is learned; reuse asks whether new objects or expressions can use it; retention asks whether it remains available after further learning. The practical variables are visible information, supervision targets, and preservation conditions. The principle developed through model construction and mechanism experiments before informing the third-stage design.

%% file: tables/leaderboard.tex
Qiushi Engine / Principle-guided & 42.25 & 53.15 & 4.10 \\
Qiushi Engine / Frontier & 42.02 & 52.86 & 4.08 \\
go76dof / WWM curriculum & 41.80 & 52.97 & 2.71 \\
Serdar404 / RecGPT-10M & 41.53 & 52.40 & 3.46 \\
Wector1 / Wordpiece-24-4 & 41.31 & 52.85 & 0.92 \\
gyaudhw / compagg relay & 40.94 & 51.81 & 2.92 \\
gyaudhw / compagg relay (s0) & 40.93 & 51.79 & 2.92 \\
svsatheesh / BabySteps & 40.86 & 53.59 & -3.72 \\
qyxu1994 / instanton-hybrid & 40.78 & 51.53 & 3.18 \\
gyaudhw / synonly (s0) & 40.67 & 51.46 & 2.91 \\
\bottomrule

%% file: chapters/03_frontier.tex
\chapterstart{Stage I: Frontier Advancement}
\label{sec:frontier}

Building a frontier model under strict data limits required coordinated choices about text, architecture, and training. Qiushi Engine compared data processing and model designs, then investigated how to improve a mature model as its exposure budget approached the limit. Complete evaluation determined which combinations succeeded. Beyond the resulting model, this work established the experimental basis for continued training and mechanism analysis.

\subsection{Choosing Representations, Architectures, and Objectives}

Text representation determines the units a model learns from. Character representations produce longer sequences of finer units; subword representations combine frequent fragments and require embeddings for a larger vocabulary. Autoregressive models predict subsequent text from preceding text, while masked language models recover hidden tokens from visible context. Qiushi Engine compared these choices along with depth, width, vocabulary, and training schedules.

Other candidates used sparse computation, auxiliary memory, or additional word-form features. Lower loss on one objective was informative but insufficient: a configuration still had to improve the relevant language abilities under comparable budgets. Architecture comparisons therefore required shared-parameter checks; tokenizer comparisons required revised sequence and exposure accounting; objective comparisons required measuring both gains and costs. The investigation moved from local component effects to a training design whose parts worked together.

The final route combined compact restatements, residual learning, and evaluation-based selection of a late-training increment. Other architectural studies retain their independent results in Section~\ref{sec:lineage} and Appendix~\ref{app:catalog}. The following sections describe the actual formation of the released model.

\subsection{Compact Restatements and Budget Reinvestment}

Original passages are paired with corresponding rewrites. An original passage is the \emph{source}; an alternative expression of its content is the \emph{restatement}. Experimental records also use \emph{text view} and \emph{aligned paraphrase} for this second expression. A source--restatement pair creates a prediction problem different from exact repetition, but both expressions consume words. Compact restatements reduce redundancy while retaining their connection to the source, freeing budget for additional material.

Shortening each restatement released words for additional passage pairs. The core collection grew from 10,094 to 12,155 pairs, an increase of 2,061 or 20.42\%, within a fixed-size corpus subset called the replacement block. Its 423,520 words comprise 261,803 source words, 161,708 restatement words, and nine padding words. The pairs came from 4,529 source documents, each of which could supply multiple passages.

\begin{table}[htbp]
\caption{Compact restatements expand paired experience within a fixed word budget. The replacement block remains at 423,520 words.}
\label{tab:compact}
\small
\begin{tabularx}{\linewidth}{@{}Y r Y@{}}
\toprule Quantity & Count & Interpretation\\\midrule
Original core pairs & 10,094 & Initial paired experience\\
Additional pairs & 2,061 & Reinvestment of saved words\\
Final pairs & 12,155 & 20.42\% more than the original set\\
Source documents & 4,529 & May contribute multiple pairs\\
Source/restatement words & 261,803/161,708 & Both expressions retained\\
Replacement-block words & 423,520 & Includes nine padding words\\
\bottomrule
\end{tabularx}
\end{table}

Compression and reinvestment change both expression and coverage: the restatement becomes shorter, and additional pairs broaden the available source information. Distinguishing these effects subsequently motivated fixed-budget substitution and target-deletion studies.

Training revealed a reversal in relative performance. A seven-metric screening average combined BLiMP, Supplement, EWoK, Entity Tracking, COMPS, GlobalPIQA, and Reading. Because it includes Reading and excludes (Super)GLUE, it is not the official NLP Average. In the corresponding matched comparison, the compact-restatement method was about 0.83 points below its reference at 20M cumulative words, but 1.29 and 1.35 points above at 70M and 80M. An early ranking would therefore have missed the benefit that emerged later, motivating study of data value across learning trajectories.

The stage also investigated continuous window packing, which organizes adjacent text into sequences the model can process. These studies produced methods for word-budget use, truncation, and context organization. Only the adopted data and weights entered the final model's lineage; the remaining implementations are retained as separate research outputs.

\subsection{Residual Structure and Continued Incremental Learning}

The representative model uses a DeBERTa-v2-style masked language architecture~\citep{he2021deberta,microsoft2021debertav2}: eight layers, hidden width 480, eight attention heads, and a vocabulary of 16,384. Hidden width is the size of the internal vector at each position; attention heads extract information from different contextual positions. Pretraining uses a 256-token prediction window. A byte-level BPE tokenizer, trained on the restricted corpus, merges frequent byte sequences into vocabulary units. The ordinary objective uses 15\% whole-word masking, masking a selected word's subword tokens together.

A residual branch adds a learned correction to an existing representation. Here the branch projects width 480 down to 128 and then back to the original width. During joint backbone training, its form is
\begin{equation}
h'=h+s\,U\,\sigma\!\left(D\,\operatorname{Norm}(h)\right),
\label{eq:residual}
\end{equation}
where $D$ and $U$ reduce and restore dimensionality, $\sigma$ is a nonlinear transformation, $\operatorname{Norm}$ normalizes the representation, and $s$ scales the correction. The upward projection is initialized to zero, so the new branch initially leaves the representation unchanged. Joint training uses $s=1.75$, with the relevant parameters learned together from random initialization. The construction draws on residual and parameter-efficient learning~\citep{bachlechner2021rezero,houlsby2019adapters}; here it is combined with budget-constrained data organization, training from scratch, and later incremental learning.

At 82,012,495 words of cumulative exposure, the jointly trained model has 35,463,008 parameters. Qiushi Engine then froze those parameters and added a dedicated trainable increment containing 995,584 parameters in 48 tensors. Total model size became 36,458,592 parameters. This stage used 101 updates and 3,992,800 additional words, bringing cumulative exposure to 86,005,295.

\begin{table}[htbp]
\caption{The two training phases in the first-generation model's lineage. Trainable parameter counts and computational savings are separate quantities.}
\label{tab:lineage}
\small
\begin{tabularx}{\linewidth}{@{}Y Y r r@{}}
\toprule Phase & Updated parameters & Total parameters & Cumulative words\\\midrule
Joint backbone training & Backbone and residual branches & 35,463,008 & 82,012,495\\
Frozen-base incremental training & New branch only & 36,458,592 & 86,005,295\\
\bottomrule
\end{tabularx}
\end{table}

Incremental training used both ordinary prediction loss and a KL preservation term. Kullback--Leibler divergence measures the difference between predictive probability distributions; the preservation loss constrains changes to the previous model's predictions. Thus, the first generation already used output preservation. The subsequent question was more specific: under which inputs should which functions be preserved, and how should this constraint coexist with new learning?

The dedicated increment provides a useful control. Disabling it restores the frozen base's output; enabling it adds the learned correction. Retention is evaluated with the branch enabled. Only about 2.73\% of parameters are trainable. Freezing the base removes its parameter updates, but gradients must still pass through intervening frozen operations to train incremental modules at earlier layers. Computation also includes reference-model predictions. The trainable parameter fraction is therefore not a compute-saving estimate.

\subsection{Coherent Relations in Late Training}

Freezing a common base permits direct comparisons of text organization. A key control held the starting model, increment, exposure, and update count fixed while shuffling coherent segments within each training row. At residual scale 1.0, coherent training achieved 44.1064 on the seven-metric average, compared with 43.1214 after segment shuffling: a difference of approximately 0.9850. The same words produced different outcomes when their coherent organization changed.

\begin{table}[htbp]
\caption{Late-training references and controls. The screening average comprises BLiMP, Supplement, EWoK, Entity, COMPS, GlobalPIQA, and Reading. The final two rows are the closest organization-only comparison; ordinary and incremental continuation also differ in updated parameters, optimizer, and objectives.}
\label{tab:late}
\small
\begin{tabularx}{\linewidth}{@{}Yrr@{}}
\toprule Condition & Cumulative words & Seven-metric mean\\\midrule
Backbone before continuation & 82,012,495 & 43.9594\\
Ordinary joint continuation & 86,006,729 & 43.7707\\
Increment with mismatched correspondence & 86,005,413 & 44.0129\\
Coherent increment, scale 1.0 & 86,005,295 & 44.1064\\
Increment with within-row segment shuffling & 86,005,295 & 43.1214\\
\bottomrule
\end{tabularx}
\end{table}

After training the increment at scale 1.0, evaluation of alternative inference scales selected 0.75 for release: three quarters of the learned branch output is added to the original representation. This was an evaluation-based model-selection decision. The final first-generation local Overall was 42.0240, displayed publicly as 42.02.

Individual answers and local capabilities also changed as aggregate performance improved. These observations led directly to the next questions: how does incremental learning balance new functions with established abilities, and which inputs can still use the computation that has been learned?

\subsection{The Experimental Basis for Principle Discovery}

Stage I produced a model that could be trained further, reusable source--restatement data, and phenomena requiring explanation. Coherent organization outperformed segment shuffling; compact-restatement benefits changed over training; local improvements did not always yield aggregate gains. Qiushi Engine used these observations to move from configuration selection to mechanism studies of learned relations, reuse on new inputs, and retention during further learning.

%% file: chapters/04_principles.tex
\chapterstart{Stage II: Principle Discovery}
\label{sec:principles}

Stage I showed that changing text organization and training could improve a model. In Stage II, Qiushi Engine turned those practical results into questions about experience, representation, objectives, architecture, reusable capabilities, learning dynamics, and measurement. It proposed explanations, designed controls that could distinguish them, conducted experiments, and revised its understanding. The outputs included both principles relevant to the next model and independent theoretical constructions, mechanism studies, and experimental methods.

Two complementary investigations connect experience to capability: natural-text experiments establish how training changes context use, while controlled tasks test whether new inputs retain access to learned computations after further training. Their findings inform the next training design. Section~\ref{sec:lineage} develops the other research branches, and Appendix~\ref{app:catalog} records their methods and evidential status.

\subsection{Relation Type Shapes Context Use}
\label{sec:relation}

Exact repetition presents the same expression again. Aligned restatement presents corresponding content in different words. Both increase exposure to related text, but impose different prediction problems. If they supplied interchangeable information exposure, comparable budgets should produce similar source-use behavior. If the model instead practices particular source--target relations, transfer should depend on the relation required by the later target.

Consider a source that describes an object's location and a restatement that describes the same location in another sentence form. Predicting the masked location could require reading the source or merely recognizing a local collocation. Replacing the source with length-matched unrelated text while leaving the target unchanged measures the contribution of that source. We use \emph{contextual dependency} to mean the information relationship needed for the target prediction.

All conditions inherit the existing rewritten corpus. The reference adds none of the compact pairs under investigation; two other conditions add exact repetition or aligned restatement. These test the relation in the additional experience. Each pairing condition also has same-window and split-window variants. The latter preserve that condition's text but place the source and corresponding passage in separate windows, so a prediction cannot access both. Thus, relation-type comparisons and within-condition, text-matched window comparisons test different variables.

Source use is measured with fixed target tokens under true-source, unrelated-source, and length-matched neutral-context inputs. For a correct token with probability $p$, negative log-likelihood (NLL) is $-\log p$: smaller values mean better predictions. Averaging over the specified targets gives cross-entropy (CE); natural logarithms express the result in nats. Let $T,U,N$ denote the respective mean losses. Define
\begin{equation}
A_T=N-T,\qquad A_U=N-U.
\label{eq:source_advantage}
\end{equation}
Larger $A_T$ means that the true source helps more than the neutral context. $A_U$ tests whether unrelated source material produces a similar change. A substantial improvement in $A_T$ with little change in $A_U$ supports an effect specific to the relevant source content.

\begin{figure}[htbp]
\centering\makebox[\linewidth][c]{\includegraphics[width=\ReportFigureWidth]{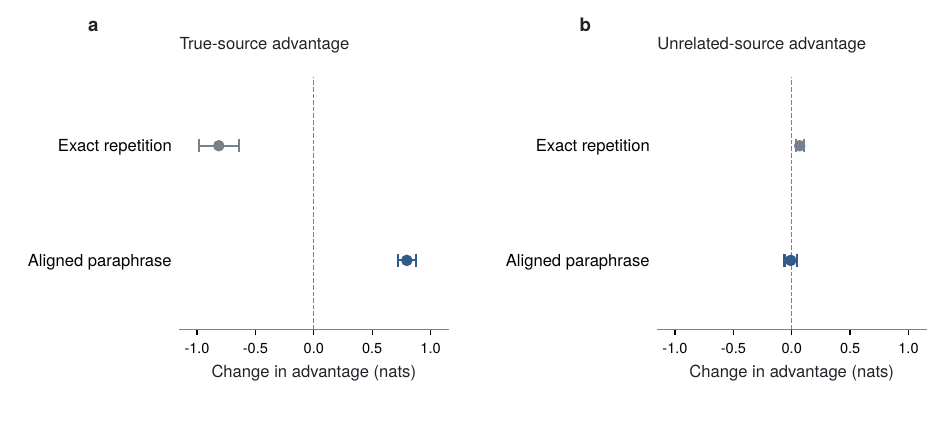}}
\caption{Training relations change the use of source information. (a) True-source advantage; (b) unrelated-source advantage. The horizontal axis shows the change from the reference model in nats, as defined in Equation~\eqref{eq:source_advantage}. Targets are compact-restatement tokens absent from the source by tokenizer ID. Points are means over three training seeds; error bars are between-seed standard deviations.}
\label{fig:relation}
\end{figure}

Targets are also grouped by whether their exact tokenizer ID occurs in the source. This measures lexical token overlap, not whether a whole word, fact, or meaning is new. Figure~\ref{fig:relation} uses nonoverlapping target tokens in compact restatements.

Relative to the reference, exact repetition changes true-source advantage by $-0.8138$ nats and aligned restatement by $+0.7970$ nats, averaged over three seeds. Standard deviations are 0.1729 and 0.0758. Unrelated-source advantage changes much less: approximately $+0.0679$ and $-0.0085$ nats.

Does this difference depend on the text alone, or on the opportunity to use its parts together? Table~\ref{tab:relation_windows} reports window controls for two seeds. Separating the two passages, while retaining each condition's complete material, substantially attenuates both the negative repetition effect and the positive restatement effect. The text remains in training but no longer supplies a shared context for a single prediction. Organizing limited experience therefore requires deciding not only which passages to include, but which relationships a prediction can access.

\begin{table}[htbp]
\caption{Window controls preserving each condition's paired text. Values are changes in true-source advantage, $\Delta A_T$ (nats), averaged over the 80M-, 90M-, and 100M-word checkpoints. Each checkpoint is measured on 2,732 source-nonoverlapping targets. Columns correspond to different training seeds.}
\label{tab:relation_windows}
\small
\begin{tabularx}{\linewidth}{@{}Ylrr@{}}
\toprule Training relation & Window layout & Seed 43022 & Seed 43122\\\midrule
\input{tables/relation_windows}
\end{tabularx}
\end{table}

Natural restatements from English and Simple English Wikipedia further reveal target dependence. Table~\ref{tab:natural_restatement} compares 1,200 source pairs separately for targets whose token IDs occur in the source and targets whose IDs do not. Aligned restatement improves true-source advantage when a source token recurs in a different sentence form. Without token overlap, its effect is close to the reference, whereas exact repetition produces a negative effect. Reusing source content in another expression and predicting a source-absent target impose different requirements; training does not transfer indiscriminately between them.

\begin{table}[htbp]
\caption{Transfer to natural restatements. Entries are changes in true-source advantage, $\Delta A_T$ (nats). Targets are averaged within each source pair and target class, then across pairs. Values are means and standard deviations over three training seeds. Each target class contains 1,200 source pairs; overlap is defined by tokenizer ID.}
\label{tab:natural_restatement}
\small
\begin{tabularx}{\linewidth}{@{}Yrr@{}}
\toprule Target class & Exact repetition & Aligned restatement\\\midrule
\input{tables/natural_restatement}
\end{tabularx}
\end{table}

Transfer is selective: its direction depends on how the practiced relationship corresponds to the relationship required at test time. Same-window versus split-window comparisons identify when that relationship can enter learning; natural-restatement tests identify which targets benefit. Together, these results turn the broad idea of adding related text into a testable principle of experience organization.

This finding connects to work on related-document pretraining, training distributions, and contextual learning~\citep{shi2024incontext,chan2022distribution,haga2024variationsets}. Chen et al. directly studied within-window parallel structure by removing it from pretraining text~\citep{chen2024parallel}; repetition can also support the formation of particular attention mechanisms~\citep{zucchet2025sparse}. The present experiments distinguish specific effects of repetition and restatement across target classes, test window separation while retaining each condition's text, and identify transfer settings without the same benefit.

\subsection{Familiar Performance and Reuse on Unseen Inputs}
\label{sec:reach}

After a model learns a relation, can new names or symbols use the same computation? Its \emph{functional reach} is the range of inputs able to use that computation; \emph{functional accessibility} describes whether a particular input can do so. For example, selecting an attribute for a familiar name and performing the same operation for an unseen name are separate tests. Qiushi Engine studied a four-choice relation task with familiar and unseen symbols and intervened directly on internal signals involved in answer selection.

The task specifies a query object and four object--attribute assignments, then asks for the queried object's attribute. An illustrative instance is ``query B; A is red, B is blue, C is green, D is yellow,'' whose answer is blue. The experiments use controlled symbols rather than these explanatory words. Tests can retain the rule while replacing the query symbol with one absent from the corresponding relation training. They can also query familiar objects within contexts containing unseen objects. The latter separates difficulty with unfamiliar context from difficulty selecting an attribute using an unfamiliar query.

\begin{figure}[htbp]
\centering\makebox[\linewidth][c]{\includegraphics[width=\ReportFigureWidth]{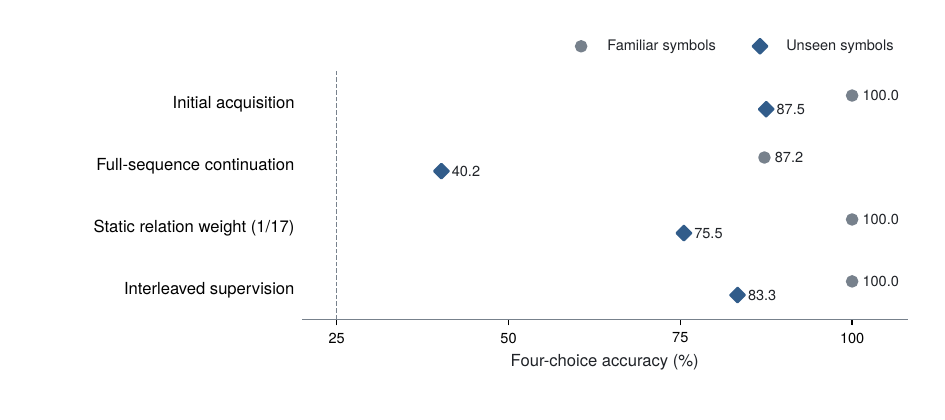}}
\caption{Four-choice accuracy on familiar and unseen symbols. Conditions are initial acquisition, full-sequence continuation, static relation supervision, and interleaved supervision. Points average two training seeds; the dashed line denotes 25\% chance accuracy. The static relation weight is $1/17$; interleaving alternates relation-answer and full-sequence targets.}
\label{fig:reach}
\end{figure}

Full-sequence supervision predicts sequence targets; relation-answer supervision emphasizes the query's answer position. A static condition fixes the relative weight of these signals, while interleaving alternates relation-answer and full-sequence supervision. Figure~\ref{fig:reach} reports means for the two training seeds used in the functional-intervention study, 43 and 100, both of which had acquired unseen-symbol transfer during initial relation learning. Initial accuracy is 100.0\% on familiar symbols and 87.5\% on unseen symbols. Full-sequence continuation yields 87.2\% and 40.2\%. With a static relation weight, unseen accuracy is 75.5\%; with interleaved supervision it is 83.3\%, both at 100.0\% familiar accuracy. Supervision allocation changes whether new symbols can use the learned relation.

Accuracy alone does not identify the computation producing an answer. Each attribute position has an internal vector, or hidden state. In the first layer, the study takes the normalized difference between the mean hidden states at queried and other attribute positions. This selection-related direction is fitted on 1,000 examples from the initial relation model and then held fixed across continuation conditions; each test class uses 384 examples. Interventions change only the projection onto this direction. Centering sets it to the same value at all four positions; zeroing removes it; rotation cyclically transfers its values between positions. All orthogonal components remain unchanged. If removing the signal impairs the answer and moving it redirects the answer, the intervention establishes a functional role in selection.

Zeroing reduces initial unseen accuracy from 87.5\% to 25.0\%, and interleaved-supervision accuracy from 83.3\% to 38.7\%. After rotation, a separate measurement asks whether the answer follows the relocated signal. This intervention-target selection rate is distinct from accuracy on the original question.

\begin{table}[htbp]
\caption{Functional interventions on unseen symbols, in percent, using the same setting as Figure~\ref{fig:reach}. The first three numerical columns measure original-question accuracy. The final column measures selection of the new target indicated by the rotated signal.}
\label{tab:intervention}
\small
\begin{tabularx}{\linewidth}{@{}Yrrrr@{}}
\toprule Condition & Intact & Centered & Zeroed & Relocated target\\\midrule
Initial acquisition & 87.5 & 26.6 & 25.0 & 67.6\\
Full-sequence continuation & 40.2 & 25.1 & 31.1 & 31.4\\
Static relation weight & 75.5 & 31.3 & 33.6 & 48.2\\
Interleaved supervision & 83.3 & 32.9 & 38.7 & 63.9\\
\bottomrule
\end{tabularx}
\end{table}

Measurements through continuation separate loss of the signal from later recovery. Full-sequence training first weakens the answer-selection signal, then restores its use for familiar queries, without comparable recovery for unseen queries. Familiar queries still use the signal in contexts containing unfamiliar objects. The distinction concerns which queries can access a learned computation, rather than a general inability to process unfamiliar context.

An alternative explanation is that the relevant information moves to another linear direction after continuation. The study therefore fits new directions from familiar queries, a single unseen query, or two unseen queries, keeping fitting and testing examples separate. It replaces answer-selection signals along each new direction and measures whether answers move to the designated position. Refitting alone does not alter the model's predictions; the fitted direction is an intervention tool. Unseen-query redirection remains weaker after full-sequence continuation than after interleaving. These tests do not support the explanation that changing the linear direction suffices to recover the same function. They connect information that can be read from a representation with computation that actually affects behavior~\citep{geiger2021causal,zhang2024patching}.

The loss of contextual learning during training and differences between familiar and unseen tokens have prior evidence~\citep{singh2023transient,anand2025dual}. Dual Process Learning studies the coexistence of these strategies through weight forgetting. Here the new measurements concern the effects of relation-supervision allocation and the functional consequences of removing or moving a specific signal.

\subsection{How Supervision Allocation Changes Relation Learning}

Visible source text is only one prerequisite for relation learning. If nearby words already determine the answer, loss can fall without much source use. Supervision allocation determines which predictions are trained and their relative weight. Removing local clues, changing target positions, or changing loss weights can alter the learning problem; relation behavior and ordinary language performance must both be measured.

The controlled experiments compare full-sequence, relation-answer, interleaved, and effective-weight-matched supervision. Relation-misaligned answer targets and a simple reduction in context pressure do not reproduce all the target behavior. The correspondence between the supervised position and the relation being learned is therefore an experimental variable, not merely a matter of total loss or the amount of relational text.

Target-deletion experiments with compact restatements test a related question while preserving the input. A source-absent target is a word absent from its paired source passage; a copied target has a word-level match there. The study removes some prediction losses from either class. At 100M words, the seven-metric average is 43.896 with full supervision, 43.437 after deleting source-absent targets, and 42.792 after deleting matched copied targets. Target selection changes later performance even with the same input. The deletion counts are close but unequal, while full supervision has more targets, so content and quantity are both involved.

A separate measurement uses the target-deletion models at 20M words and excludes source pairs used in the corresponding training. Across 710 pairs and 974 target events, deleting source-absent supervision rather than matched copied targets raises measured source-absent loss by 0.0796 nats. Resampling by source pair gives a 95\% interval of $[0.031,0.125]$. Requiring whole-document separation yields an interval containing zero. This early local-prediction test and the 100M-word task evaluation answer different questions: which targets support prediction of specified content, and how the supervision choice affects later task performance. Appendix~\ref{app:extended} preserves both measurements.

Together, these experiments motivate separate tests of whether input supplies the relation, targets require its use, and sufficient learning signal reaches those targets. Separating input corruption from prediction rate has direct precedent~\citep{wettig2023mask}. The present work applies that distinction to source--restatement learning and measures the consequences of selecting different supervised positions in otherwise retained text.

\subsection{Experience Value Depends on Budget, Displacement, and Learning State}

Adding data under a fixed budget necessarily displaces other material. Replacing an equal amount of experience estimates a substitution effect. Let $A$ be added experience, $D$ displaced experience, and $B$ the reference stream. Under fixed training settings, define
\begin{equation}
\Delta_T(A,D;B)=Y_T(B-D+A)-Y_T(B),
\label{eq:substitution}
\end{equation}
where $Y_T$ is the evaluation at training amount $T$. The comparison holds budget, architecture, schedule, and seed fixed. This is an experimental quantity to estimate, not a formula that predicts data value without measurements.

Late-training DeBERTa comparisons distinguish transformed views, expanded sentence coverage, and exact repetition. In another RoBERTa configuration, the view method improves its own training objective without yielding the same broad benefit. Holding the added content fixed while removing child-directed or adult-prose material also changes the result. Experience value depends on displaced information, learning stage, and model state; one early loss or text-distribution distance does not establish a universal ranking.

These studies also uncovered an initialization confound. A seed determines a pseudorandom sequence, but inserting a layer changes how its values are assigned to parameters. The same seed can therefore give shared tensors different initial values. Explicitly copying the common parameters substantially weakened a previously large negative interaction, where interaction means that a data treatment has different effects in different architectures. Actual parameter alignment is essential to that comparison.

\subsection{A Design Principle and Independent Findings}

Stage II made three conditions separately testable: the information visible during prediction, the targets and weights that receive supervision, and the existing functions to preserve during further training. Natural-text experiments establish effects of relation and window organization; controlled tasks establish conditions for reuse on new inputs. They jointly motivate coordinated input, supervision, and preservation design.

The corresponding tests are concrete: hold targets fixed while increasing local masking; hold masking fixed while adding targets; change the inputs used for preservation while measuring both acquisition and retained behavior. Stage III implements these choices in real language-model training and uses its controls to refine the design.

Relational anchors, shared representation spaces, identity matching, entity storage, and measurement design also retain independent value. Section~\ref{sec:lineage} explains their questions, constructions, and results. Their scientific value does not depend on inclusion in the second-generation training recipe.

%% file: tables/relation_windows.tex
Exact repetition & Same window & $-0.683$ & $-1.010$ \\
Exact repetition & Split windows & $+0.009$ & $-0.134$ \\
Aligned restatement & Same window & $+0.712$ & $+0.824$ \\
Aligned restatement & Split windows & $+0.116$ & $+0.028$ \\
\bottomrule

%% file: tables/natural_restatement.tex
Target token present in source & $+0.109\pm 0.122$ & $+0.238\pm 0.071$ \\
Target token absent from source & $-0.217\pm 0.029$ & $+0.032\pm 0.040$ \\
\bottomrule

%% file: chapters/05_guided_improvement.tex
\chapterstart{Stage III: Principle-Guided Model Improvement}
\label{sec:guided}

Qiushi Engine returned to the first-generation model with two findings: presenting text does not ensure use of its relationships, and acquiring new behavior does not ensure retention of existing functions. It retained the \Frontier{} weights, tokenizer, residual structure, and existing text while changing the training task: which clues remain visible, which tokens must be predicted, and how changes to established predictions are constrained. Comparison with ordinary continuation then tested the practical value of the design.

\subsection{From Mechanism Findings to Training Decisions}

Relation experiments showed that including the source within the prediction window changes learned source use. Supervision and internal interventions further showed that available information can go unused, and familiar performance need not establish reuse. These findings led to three controllable design choices (Table~\ref{tab:bridge}).

\begin{table}[htbp]
\caption{Connections between scientific findings and the new training design. Each finding motivates a specific intervention and a corresponding comparison.}
\label{tab:bridge}
\small
\begin{tabularx}{\linewidth}{@{}Y Y Y@{}}
\toprule Finding & Training intervention & Direct test\\\midrule
A visible source need not be used & Retain the source; remove more local clues in the restatement & Hold targets fixed and vary masking; measure source-condition effects\\
Input conditions and target counts are separate variables & Use dense masking with sparse supervised targets & Compare sparse and dense targets; complete model evaluation\\
Preservation can also suppress new learning & Preserve parent behavior on ordinary inputs & Compare acquisition, prediction changes, and task scores under ordinary and dense-input preservation\\
\bottomrule
\end{tabularx}
\end{table}

The design developed through revisions. Early explanations emphasized the number of prediction targets. Effects persisted when sparse targets were fixed and masking increased, directing attention to information availability separately from supervision quantity. After developing ordinary-input preservation, the study compared alternative preservation inputs and update magnitudes. These were iterative hypotheses and tests, rather than one complete method fixed before all experiments.

\subsection{Separating Visible Inputs from Supervised Targets}

Source--restatement pairs come from the first-stage data; no new rewriting is performed in this continuation. Ordinary whole-word masking may leave enough local clues to predict a restatement target without consulting the source. The new method retains the source and masks more candidate words in the restatement, while computing focused prediction loss at only a selected subset. Character spans map candidate words to one or more tokens; these jointly treated units are called \emph{content groups}.

Masking determines what the model can see; supervision determines which prediction errors update its parameters. A position may be masked without contributing its own loss, removing a clue for another target. Dense masking with sparse supervision expands the missing local context without requiring prediction at every masked position.

\begin{figure}[htbp]
\centering\makebox[\linewidth][c]{\includegraphics[width=\ReportFigureWidth]{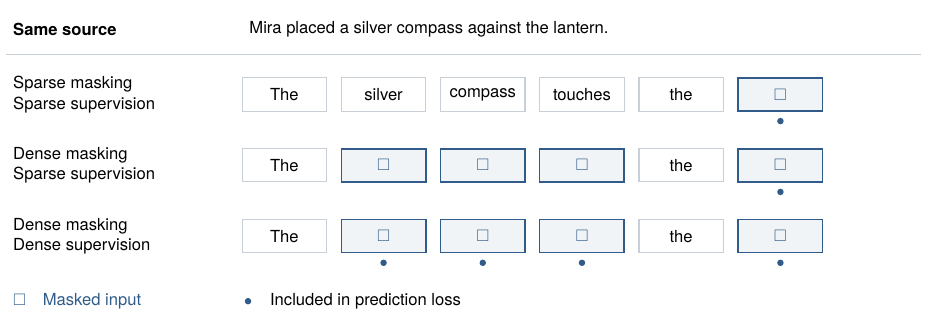}}
\caption{Separating input masking from supervision, illustrated at word level. The first two rows retain targets while changing masking; the final two retain masking while changing targets. Empty boxes denote masked words; dots mark positions contributing prediction loss. Candidate content words are selected while stopwords remain visible. Appendix~\ref{app:repro} gives the actual selection and tokenization rules.}
\label{fig:mask}
\end{figure}

Let $(I,T)$ denote the combination of input masking and focused targets, with $S$ for sparse and $M$ for dense. \emph{Focused targets} are positions selected specifically for source--restatement prediction; their loss is the focused loss. $(S,S)$ and $(M,S)$ share sparse focused targets but change visible clues. $(M,S)$ and $(M,M)$ share dense masking but change which positions contribute focused loss (Figure~\ref{fig:mask}).

The continuation stream contains 20,475 packed rows and 3,162,742 words. A row is a training input that may contain several segments. The focused relation objective applies to 3,831 rows of existing Qwen rewrites, containing 11,778 paired segments. Other rows retain ordinary language prediction, including Stage I's compact-restatement and budget-reinvestment material. Dense masking covers 132,283 content groups and 176,607 tokens; sparse supervision selects 21,479 groups and 28,590 tokens. Thus, 148,017 tokens are masked without contributing the focused loss in $(M,S)$. Sparse and dense are defined relative to the candidate set; Appendix~\ref{app:repro} gives the exact rules.

The acquisition objective combines relation prediction with ordinary language prediction:
\begin{equation}
\mathcal L_{\mathrm{acq}}
=0.15\,\overline{\CE}_{\mathrm{focus}}
+0.85\,\overline{\CE}_{\mathrm{ordinary}}.
\label{eq:acquisition}
\end{equation}
For each optimizer update, the focused and ordinary losses are each summed over all valid targets in their respective rows, then divided by their own total target-token counts before weighting. This is not an equal-weight average of row or microbatch means. The coefficients weight the two learning signals; they are not fractions of text or tokens. The ordinary objective continues broader language learning while the focused objective trains the relation task.

Input corruption and prediction supervision have been separated in prior work~\citep{wettig2023mask}. This study uses that distinction for source--restatement learning: keep the source available, alter local clues on the restatement side, and choose supervised targets separately. The comparisons test both the behavioral effect and the benefit to the first-generation model.

\subsection{Preserving Functions on Ordinary Inputs}

The new task permits changes to predictions when local clues are missing, while preservation protects behavior under ordinary input conditions. The same existing Qwen paired rows used for relation learning are presented again with ordinary 15\% whole-word masking. The updated model and frozen \Frontier{} predict the same targets from the same complete rows, with sources retained. KL divergence constrains differences between their probability distributions. Here \emph{ordinary input} denotes the masking procedure, not a separate corpus.

The full objective is
\begin{align}
\mathcal L_{\mathrm{pres}}
&=\frac{1}{|Q_p|}\sum_{(x,i)\in Q_p}
\KL\!\left(p_{\Frontier}(\cdot\mid x,i)\,\|\,p_\theta(\cdot\mid x,i)\right),\\
\mathcal L&=\mathcal L_{\mathrm{acq}}+\lambda\mathcal L_{\mathrm{pres}},
\qquad\lambda=1,\quad T=1.
\label{eq:preservation}
\end{align}
Here $x$ is a complete row after ordinary whole-word masking, $i$ is a selected ordinary prediction position, and $Q_p$ collects all preservation targets in the update. KL is computed over the full vocabulary at those positions and averaged over their total count, not over all visible positions. The frozen first-generation model supplies $p_{\Frontier}$; $p_\theta$ is the updated model. The coefficient $\lambda$ sets the preservation weight. Temperature $T$ scales logits before probabilities are computed; $T=1$ leaves them unscaled. In teacher--student terminology, the teacher is the first-generation model itself.

Forward-pass randomness must be controlled. Dropout randomly disables parts of a neural network during training, so two forward passes can differ even at identical weights. Preservation uses deterministic computation, saving and restoring random-number-generator state so its extra passes do not perturb acquisition. This separates prediction changes due to updated parameters from those introduced by the comparison procedure.

Output-based preservation is directly related to Learning without Forgetting~\citep{li2016learning}. It also extends the first generation's existing KL term: the question here is how preservation conditions allow new learning while limiting unnecessary changes on ordinary inputs.

Only the existing increment's 995,584 parameters are updated, at residual scale 0.75; total parameter count is unchanged. The full method uses 80 updates and adds 517,332 preservation-word presentations beyond acquisition:
\begin{equation}
86{,}005{,}295+3{,}162{,}742+517{,}332=89{,}685{,}369.
\label{eq:budget}
\end{equation}
Controls without preservation total 89,168,037 words. Additional preservation presentations equal 16.36\% of this continuation's acquisition words and require extra forward computation. Both remain under the 100M limit. The equal-exposure acquisition comparison and the full method's combined benefit are reported separately.

\subsection{Complete Evaluation Across Two Continuation Seeds}
\label{sec:complete_results}

Table~\ref{tab:strategies} reports the common parent and four strategies evaluated on all nine metrics. Seeds 62064 and 62065 vary continuation randomness from the same pretrained model. Ordinary continuation tests whether the new design improves on continued training itself. The $(S,S)$ condition serves a different purpose: mechanism comparisons of input conditions with sparse targets fixed.

\begin{table}[htbp]
\caption{Complete Overall scores for the shared parent and four continuation strategies. The two seeds share a pretrained starting model. The full method includes additional preservation presentations; all nine component scores are in Appendix~\ref{app:local}.}
\label{tab:strategies}
\small
\begin{tabularx}{\linewidth}{@{}Yrrr@{}}
\toprule Strategy & Seed 62064 & Seed 62065 & Cumulative words\\\midrule
\input{tables/strategies}
\end{tabularx}
\end{table}

\begin{figure}[htbp]
\centering\makebox[\linewidth][c]{\includegraphics[width=\ReportFigureWidth]{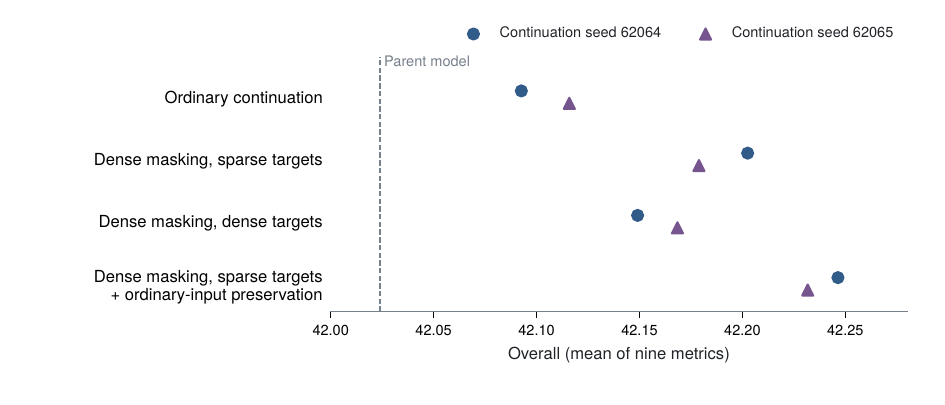}}
\caption{Complete evaluation of continuation strategies from one parent. Circles and triangles represent seeds 62064 and 62065; the dashed line is the parent's Overall. All candidates have nine-metric evaluations. The full preservation method includes additional ordinary-input presentations.}
\label{fig:strategies}
\end{figure}

$(M,S)$ exceeds ordinary continuation by 0.1099 and 0.0630 at the same cumulative word exposure. The full method exceeds ordinary continuation by 0.1538 and 0.1158, and $(M,S)$ by a further 0.0439 and 0.0529. Both dense-target $(M,M)$ models score below the corresponding sparse-target $(M,S)$ models. With the total focused-loss weight fixed at 0.15, distributing supervision over more targets does not improve Overall. This changes both the target set and the relative weight received by an individual target.

The comparisons establish an equal-exposure gain for the acquisition design and a further gain for the combined preservation method. Available $(S,S)$ results comprise screening and mechanism measurements, not the complete nine-metric table. Its masking-specific evidence must therefore be taken from those measurements; the Overall difference between $(M,S)$ and ordinary continuation does not isolate masking alone.

\begin{figure}[htbp]
\centering\makebox[\linewidth][c]{\includegraphics[width=\ReportFigureWidth]{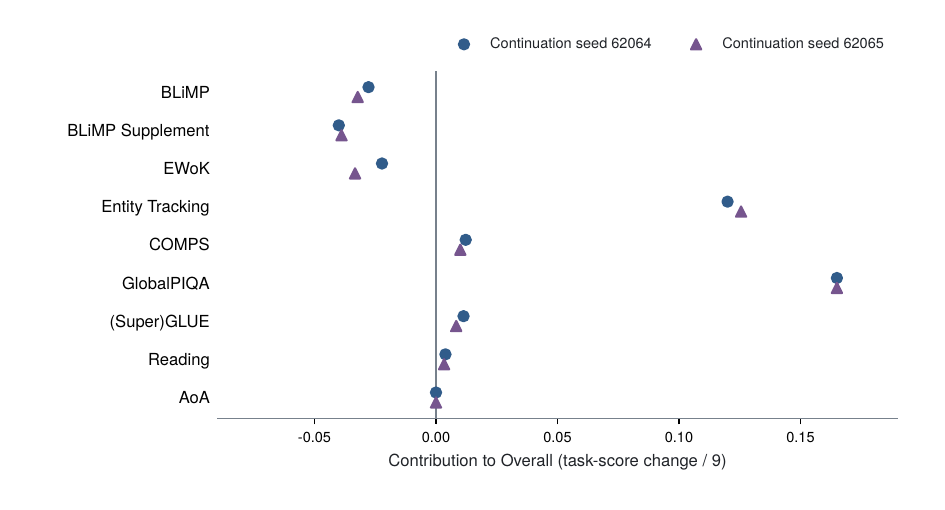}}
\caption{Component gains and costs of the full method relative to the common parent. Each task-score change is divided by nine to give its contribution to Overall. Circles and triangles denote seeds 62064 and 62065. Values are point estimates; both seeds use the same evaluation sets, including the 203-item GlobalPIQA set.}
\label{fig:contributions}
\end{figure}

Figure~\ref{fig:contributions} decomposes the full method's change from the parent. Dividing a component change by nine gives its Overall contribution. GlobalPIQA contributes about 0.165 for both seeds, or approximately 74\% and 79\% of the total gains. Entity Tracking also improves, while some components decline. The remaining eight components jointly contribute 0.0574 and 0.0428. Net improvement thus combines identifiable gains and trade-offs.

To identify what the method improves beyond continued training, ordinary continuation is the relevant reference. Against that reference, Entity Tracking supplies the largest positive Overall contribution for both seeds, $+0.1378$ and $+0.1511$. GlobalPIQA contributes $+0.0572$ in each case; other components partly offset these gains. Comparing the full method with acquisition-only $(M,S)$, GlobalPIQA is unchanged: the additional $+0.0439$ and $+0.0529$ come from the net changes in other components. Gains from the parent, advantages over ordinary continuation, and the added effect of preservation are distinct comparisons.

An additional check varies downstream fine-tuning rather than continuation. At fine-tuning seeds 42 and 44, the full method's (Super)GLUE mean exceeds both the parent and $(M,S)$. This comparison holds the continued models fixed and tests another source of randomness. Table~\ref{tab:finetuning_seeds} reports all three models under both fine-tuning seeds.

\subsection{Acquisition and Preservation Measurements}

Three quantities measure new learning on densely masked inputs: correct-target cross-entropy, the correct token's rank among vocabulary candidates, and KL from the parent to the updated model. Lower CE and a better rank indicate improved prediction; lower KL indicates less output change. The comparison uses 169 previously trained examples and 1,003 shared target tokens. It measures learned behavior on those examples, not generalization to unseen documents.

\begin{table}[htbp]
\caption{Acquisition and prediction changes on densely masked inputs. All rows use continuation seed 62064 except the explicitly labeled 62065 row. Larger CE and rank improvements relative to the parent are better; KL measures output divergence. The final row applies preservation to densely masked inputs.}
\label{tab:preservation}
\small
\begin{tabularx}{\linewidth}{@{}Yrrr@{}}
\toprule Method & CE gain (nats) & Rank gain & Parent KL (nats)\\\midrule
Ordinary continuation & $-0.005$ & 0.7 & 0.001\\
$(M,S)$ without preservation & 0.757 & 132.8 & 0.503\\
Full method, 62064 & 0.682 & 128.3 & 0.353\\
Full method, 62065 & 0.679 & 128.0 & 0.345\\
Dense-input preservation & 0.159 & 34.8 & 0.009\\
\bottomrule
\end{tabularx}
\end{table}

Ordinary-input preservation retains most of the acquisition gain. Dense-input preservation sharply reduces model change but also suppresses improvement on correct targets. A useful preservation method must limit damage to ordinary behavior while allowing the changes required by the new task. Qiushi Engine therefore compared the effective strength of the constraints under different inputs.

With the updated model, parent, and shared target positions fixed, dense masking increases parent entropy by 1.6330 nats, correct-target CE by 1.3155 nats, and teacher-to-student KL by 0.4738 nats. The preservation gradient norm on trainable parameters is 11.835 times its ordinary-input value. A gradient describes the direction and magnitude of parameter change favored by a loss. Here the two gradients have directional similarity 0.7288 but markedly different magnitudes. Equal loss coefficients do not impose equal effective constraints.

\begin{table}[htbp]
\caption{Preservation constraints at fixed models and shared target positions. The updated model is $(M,S)$ at continuation seed 62064; the reference is the first-generation parent. The first three rows give dense-minus-ordinary input differences; the final row gives the gradient-norm ratio.}
\label{tab:gradient_geometry}
\begin{tabularx}{\linewidth}{@{}Yr@{}}
\toprule Measurement & Result\\\midrule
Parent predictive entropy difference (nats) & $+1.6330$\\
Parent target CE difference (nats) & $+1.3155$\\
Teacher-to-student KL difference (nats) & $+0.4738$\\
Preservation gradient norm: dense/ordinary & $11.835$\\
\bottomrule
\end{tabularx}
\end{table}

Preservation configurations also differ in their target positions: ordinary masking has 4,762 targets; densely masked positions outside focused supervision have 6,665; their intersection contains the 1,003 positions above. Their effects therefore reflect changes in input type, target set, and effective constraint strength, whose contributions remain partly confounded. All three specify the configuration and must be considered alongside measures of new and retained behavior.

A separate ordinary-input measurement uses 384 complete existing rows and 10,348 targets, retaining sources under ordinary whole-word masking. Parent KL is 0.02449 nats without preservation and 0.00941 with the full method. CE deterioration relative to the parent falls from 0.0290 to 0.0100 nats. These measurements show reduced change on ordinary inputs; they use a different sample from Table~\ref{tab:preservation}, which measures densely masked inputs.

Source interventions use 96 pairs and 186 common targets after the continuation prefix. They are outside this continuation's trained prefix, but independence from the parent's pretraining documents has not been established. Correct-source NLL improves slightly, while wrong-source and source-removal conditions incur larger costs (Table~\ref{tab:source_final}). The full method reduces some of those costs while retaining correct-source improvement. This measures responses to source changes, separately from generalization of relational reasoning.

\begin{table}[htbp]
\caption{Source interventions on two models at continuation seed 62064. Values are target NLL differences from the parent, in nats; negative values mean improvement. Every condition uses the same targets.}
\label{tab:source_final}
\begin{tabularx}{\linewidth}{@{}Yrr@{}}
\toprule Source condition & $(M,S)$ alone & Full method\\\midrule
Correct source & $-0.024$ & $-0.022$\\
Wrong source & $+0.149$ & $+0.099$\\
Restatement only & $+0.107$ & $+0.076$\\
Current source erased from full row & $+0.210$ & $+0.127$\\
All sources erased from full row & $+0.0248$ & $+0.0101$\\
\bottomrule
\end{tabularx}
\end{table}

\subsection{How the Model Tests Refined the Principle}

Stage III established practical value and sharpened the design principle. Increasing masking and increasing target count produced different outcomes, supporting separate control of prediction conditions and supervision. Preservation had to be evaluated through both acquisition and ordinary behavior, rather than proximity to the old model alone. Source interventions further exposed correct-source benefits and wrong-source costs.

The three stages thus form a continuous development of methods: the first supplies a mature model and paired text; the second investigates conditions for information use and capability reuse; the third changes input, supervision, and preservation accordingly and obtains repeated complete-evaluation gains. Model experiments turn the findings into a usable method, while additional controls provide more precise questions for subsequent research.

The new experiments also change the criteria for the next research. Fixed-target masking controls distinguish information availability from target quantity. Gradients measured at common positions show that equal preservation coefficients can impose unequal effective constraints. Qiushi Engine retains the successful method while revising its explanation and recording more informative comparison conditions. This connection between model improvement, method development, and revised scientific understanding underlies the Research RSI discussion in Section~\ref{sec:rsi}.

%% file: tables/strategies.tex
Parent & 42.0240 & Same parent & 86,005,295 \\
Ordinary continuation & 42.0926 & 42.1159 & 89,168,037 \\
Dense masking, sparse supervision $(M,S)$ & 42.2025 & 42.1789 & 89,168,037 \\
Dense masking, dense supervision $(M,M)$ & 42.1491 & 42.1684 & 89,168,037 \\
Dense masking, sparse supervision + ordinary-input preservation & 42.2464 & 42.2317 & 89,685,369 \\
\bottomrule

%% file: chapters/06_research_lineage.tex
\chapterstart{Research Lineage and Independent Findings}
\label{sec:lineage}

The three-stage program produced both model improvements and independent scientific findings. Compact text, residual architecture, and continued training connect the two model generations through actual inheritance. Studies of relation identification, information retrieval, and measurement developed into complementary research branches. This section explains their questions, their relationship to the main investigation, and the methods and understanding they leave for subsequent work.

The studies span the path from representing text to acquiring capabilities, continuing learning, and measuring outcomes. Table~\ref{tab:research_scope} groups their contributions by scientific question. The following sections develop experience substitution, relational anchors, identity matching, and measurement controls. Appendix~\ref{app:catalog} preserves the full topic catalog and the status of its conclusions; accompanying materials provide derivations, implementations, experiments, and scientific notes.

\begin{table}[htbp]
\caption{Research questions and representative outputs across the three stages. Training methods, mechanism findings, and experimental tools each provide a basis for further research.}
\label{tab:research_scope}
\small
\begin{tabularx}{\linewidth}{@{}P{29mm}Y Y@{}}
\toprule Research area & Question & Representative outputs\\\midrule
Experience and data & How should limited words be allocated to content, expression, and related passages? & Compact restatements and reinvestment; fixed-budget substitution; relation and window controls.\\
Representation and architecture & How can sparse supervision identify a relation and make it usable by new inputs? & Character/subword comparisons; zero-initialized residual increments; anchors, shared representations, and identity matching.\\
Learning signals and capabilities & Which predictions train relation use, and which computations affect answers? & Separate input and target design; supervision allocation; familiar/unseen tests and internal interventions.\\
Optimization and dynamics & How does experience value change during learning, and how can new and old capabilities coexist? & Early/late trajectories; curriculum and optimizer comparisons; relation dosage; incremental learning and preservation.\\
Measurement and computation & Which controls make comparisons interpretable and implementations faithful? & Shared initialization; source exclusion and item-level analysis; complete model loading, exposure accounting, and gradient checks.\\
\bottomrule
\end{tabularx}
\end{table}

\subsection{Model Inheritance and the Development of Scientific Questions}

\begin{figure}[htbp]
\centering\makebox[\linewidth][c]{\includegraphics[width=\ReportFigureWidth]{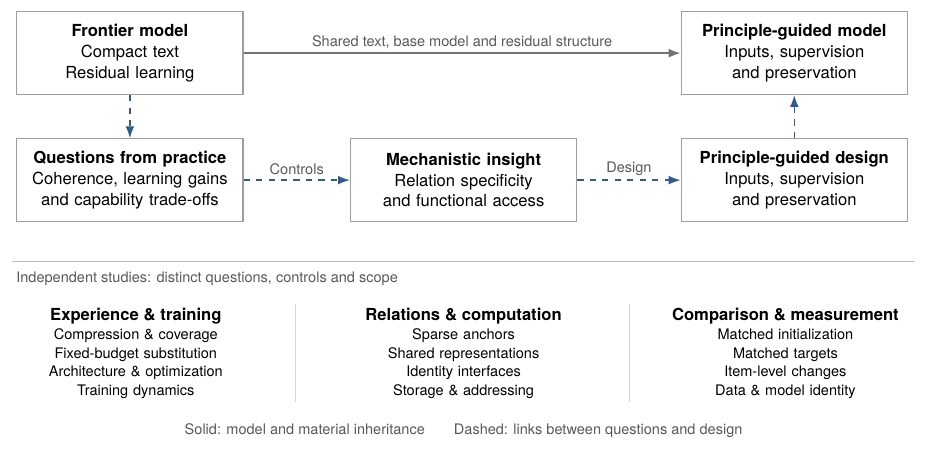}}
\caption{Model inheritance and research branches. Solid lines denote inheritance of first-generation training text, pretrained base, and residual structure. Dashed lines connect practical questions, mechanism studies, and subsequent training design. Independent findings concern experience and training, relations and computation, and comparison and measurement.}
\label{fig:lineage}
\end{figure}

Compact restatement began as a way to improve a model within its word budget; it subsequently supplied controlled material for comparing expression relationships. Residual incremental learning began as a way to continue training a mature model; it subsequently enabled comparisons of objectives and preservation with a common frozen base. A successful method can thus leave both a model and a new experimental instrument. Across Stages I and II, the objects of research remained available while explanations became more precise.

The two main mechanism branches provide complementary evidence. Natural-text experiments examine how practiced relationships shape source use. Controlled binding tasks examine how queries select attributes and whether new symbols can still use the learned computation. Different tasks and models address a shared design problem: what information a prediction can access, which targets receive supervision, and which established functions should survive further learning.

Stage III implements these choices by masking more of the restatement, retaining sparse targets, and constraining prediction changes on ordinarily masked inputs. Complete evaluation establishes the method's practical benefit. Further controls distinguish target count, input conditions, and effective constraint strength. Figure~\ref{fig:lineage} separates actual inheritance of model and data from the way mechanism findings informed the design.

\subsection{Experience Compression, Substitution, and Learning Stage}
\label{sec:experience_branch}

Data processing under a fixed budget can change the expression of existing content or allow more content into training. Compact restatement combines both. To understand their effects, the study treats what is added and what is displaced as separate experimental conditions. This fixed-budget substitution design applies to corpus construction, data mixing, and curricula, beyond the particular restatement method.

At 80M, 90M, and 100M words in one DeBERTa configuration, the study compares compact restatement, different sentences from the same source range, and near-exact repetition. Averaging late results over those checkpoints and five metrics gives differences from the reference of $+0.3853$, $+0.3350$, and $+0.0343$, respectively. The five metrics are BLiMP, Supplement, EWoK, COMPS, and Reading. The different-sentence condition broadens sentence coverage within the specified source range, rather than establishing an expansion of document coverage. Both varied expression and expanded content improve this late-training comparison, while repetition has a smaller effect.

Displaced material matters too. With an added FineWeb block held fixed, replacing child-directed/spoken material rather than adult prose produces a four-metric score 0.6125 points lower at 100M words. The respective changes from the reference are $+0.0825$ and $+0.6950$ on BLiMP, Supplement, EWoK, and COMPS. The added text is identical; different opportunity costs produce different net gains.

In a RoBERTa configuration, restatement improves its own training loss but changes the late five-metric score by $-0.6873$. The usefulness of the treatment depends on model configuration and learning stage. Text statistics, such as word-frequency distributions, can help propose candidates; their value is tested through the corresponding task measurements.

Other compression studies distinguish lexical coverage from preserved relationships. Extractive compact views, fluent connecting text with controlled vocabulary, and rule-based rewriting respectively seek broader content coverage, better connections between expressions, or reduced reliance on external generation. Short comparisons show local gains and task costs without the same completeness of final-model evaluation. Their reusable contribution is to make compression quality inspectable: which words, correspondences, and prediction clues survive, and where the released word budget is spent.

\subsection{Relational Anchors, Shared Representations, and Identifiability}
\label{sec:anchors}

Some learning problems lack enough information to determine an answer, regardless of training time. Knowing that several objects belong to the same or opposite binary classes identifies their grouping but not which group should be called positive. A few absolute labels can resolve this ambiguity. We call such labels \emph{anchors}.

The study implements this question in a binary relation task. Let an event's orientation be $z_i\in\{-1,+1\}$ and a comparison between events be $r_{ij}=z_i z_j$. A value of $+1$ denotes agreement and $-1$ opposition. For connected events, flipping every orientation leaves the pairwise comparisons unchanged:
\begin{equation}
(z_i,z_j)\mapsto(-z_i,-z_j),\qquad (-z_i)(-z_j)=z_i z_j.
\end{equation}
An absolute state label can resolve the remaining orientation ambiguity within this finite hypothesis class. Sparse anchors therefore supply missing absolute information without relabeling every relation.

The model shares input representations and sequence processing between state judgments and relational comparisons, with separate output heads. Embeddings map symbols to vectors; a gated recurrent unit (GRU) processes the sequence. Across three training seeds, reversing a few state anchors reverses indirectly supervised orientations: accuracy against the original absolute labels changes from 1.0 to 0.0. Relative same/opposite judgments remain completely correct. Absolute orientation follows the anchors while relative structure is retained.

The combination of change and invariance identifies what the anchors convey through the shared representation. A matched model with separate representations fits some local tasks but does not show the same propagation. Sharing becomes a testable condition for transferring the anchor information, rather than a general architectural preference.

\begin{table}[htbp]
\caption{Anchor reversal in the binary relation task, across three training seeds. The two measurements distinguish absolute orientation from relative structure. Accuracy is on a 0--1 scale.}
\label{tab:anchor}
\begin{tabularx}{\linewidth}{@{}Yrr@{}}
\toprule Measurement & Original anchors & Reversed anchors\\\midrule
Indirectly supervised relations, original labels & 1.0 & 0.0\\
Relative comparisons between unseen relations & 1.0 & 1.0\\
\bottomrule
\end{tabularx}
\end{table}

This construction identifies sufficient information for a specified relational structure and shows how shared representations transmit that information. It relates to work on binding objects, properties, and roles in language models~\citep{feng2024binding,dai2024binding}. Candidate objects and relation rules are provided here. Ordinary text additionally requires discovering the objects and interpreting their relations, so the one-anchor result does not directly generalize to all language learning.

Matching names to objects is another prerequisite. A model able to compute a relation still needs to associate its input name with the correct object; we call this mapping the \emph{identity interface}. Character-level experiments supply a small amount of same-letter supervision to align query and candidate representations. In the construction covering all base letters, matching accuracy rises from 0 to 1 and restores relational responses for events without direct supervision. Partial alphabet or training-name coverage leaves new names containing uncovered characters difficult.

New names in this experiment recombine learned characters. Character positions are aligned, candidate objects are provided, and the matcher is frozen after training. Hard assignment selects one candidate instead of mixing candidate representations. Under these conditions, base-character coverage supports new name combinations. Aliases, pronouns, and unrestricted reference remain distinct problems. The experiment separates learning the relation from identifying the object to which it should apply.

\subsection{Entity Storage, Retrieval, and Semantic Addressing}
\label{sec:memory_branch}

Explicit memory stores states at readable and writable locations. Language use additionally requires identifying an object in text, writing the new state to the correct location, replacing outdated content, and retrieving the state required by a query. \emph{Addressing} selects a read or write location; semantic addressing makes that selection from the meaning of the text. Qiushi Engine separates storage, retrieval, and addressing to determine what each component has learned.

Experiments first provide correct read and write locations---the gold-address condition. A slot can be understood as a numbered storage cell. Renumbering reads and writes consistently should preserve the result; exchanging only one location should change the retrieved content. These interventions distinguish invariance to arbitrary numbering from dependence on the information actually read.

With retrieved content connected to language prediction, the three-seed mean answer log-odds gain under correct addressing is 1.2056, the targeted slot-swap effect is 2.0513, and paired accuracy is 0.0472. Log-odds compares the relative probability of a target and a contrast answer; paired accuracy requires both members of a corresponding test pair to be correct. Memory affects answer probabilities, while reliable paired relational behavior remains difficult.

A subsequent design adds memory through $h+gP(m)$, where $h$ is the original representation, $m$ the retrieved content, $P$ a projection to matching dimensionality, and $g$ a gate controlling its contribution. Training also uses a lower learning rate and longer duration. Paired accuracy rises to 0.1375 and the slot-swap effect to 3.087. These are joint changes to the connection and training configuration, not an isolated estimate of the gate's contribution.

\begin{table}[htbp]
\caption{Distinct components of entity-memory behavior. Conditions test stored-content effects, address recognition, candidate selection, and temporal state selection.}
\label{tab:memory_layers}
\small
\begin{tabularx}{\linewidth}{@{}P{34mm}Y Y@{}}
\toprule Condition & Observation & Functional interpretation\\\midrule
Gold addresses & Slot swaps change probabilities; paired accuracy remains low & Memory influences prediction, but the complete relation task is difficult.\\
Addresses inferred from text & 90.4\% overall token accuracy; zero key-query recall & Easy positions conceal failure on the required queries.\\
Frozen reader, learned selector & About 0.994--0.995 on lexical and role swaps & Correspondence selection works among provided candidates.\\
Temporal state selection & About 0.511 on original/updated state selection & Determining which temporal state to retrieve remains difficult.\\
\bottomrule
\end{tabularx}
\end{table}

The natural-text difficulty motivated a selection--retrieval decomposition. A reader is first trained to retrieve content from a given location and frozen. A separate selector then uses a softmax over four candidate locations. This distinguishes inability to read the content from selection of the wrong location when the content is readable. Lexical and role swaps are largely successful, whereas choosing an original or updated state from temporal meaning remains difficult.

Task inspection also removed an apparent positive result. In an early small entity task, the correct answer did not vary with the queried object. A constant answer or first-slot shortcut could therefore produce high accuracy. That result was withdrawn as evidence for compositional binding; correcting the corpus did not itself constitute a new training validation. The storage, addressing, and selection results above come from separately completed experiments.

\subsection{Optimization, Initialization, and Measurement}
\label{sec:measurement_branch}

Architecture comparisons require a genuinely shared starting point. In an experiment removing position-related attention terms, the same random seed left 53 of 140 shared tensors with different initial values because model construction consumed random numbers in a different order. Explicit parameter copying changed the estimated architecture-by-data interaction, compact views relative to repetition, from $-0.6908$ to $+0.0679$, with an interval spanning zero. The reduced architecture still had lower absolute performance.

Two questions must therefore be distinguished: whether deleting a component reduces overall capability, and whether that component changes the relative benefit of a data treatment. The first remained supported; the initially large interaction estimate was affected by initialization. Checking tensor names, shapes, and actual starting values, then aligning common parameters explicitly, provides a reusable control.

Optimization studies likewise separated local measurements from task outcomes. They compared the directions of parameter change favored by different objectives, projected conflicting components, added detection tasks, or switched optimizers. Some interventions improved short-run loss or internal measurements without establishing stable full-task gains. More aligned updates are a hypothesis about learning, not a substitute for measuring the resulting capabilities.

Evaluation uses complementary scales. A macro-average weights task scores equally; an item-level comparison tracks whether each question changes from incorrect to correct or vice versa. Unequal task sizes allow a macro-average to rise while the total number of correct answers falls. The study records gained items, lost items, and their difference, keeping nonaccuracy quantities such as Reading separate.

Implementation choices also affect scientific interpretation. Continuous packing requires checking whole words, source--restatement correspondence, and the content actually visible after truncation. Microbatching requires weighting by target count to preserve the intended average loss. Gradient checkpointing saves memory by recomputing intermediate activations during backpropagation; short comparisons under fixed randomness test numerical consistency. Trainable parameter fraction, final-model training cost, and the cost of searching for the method are recorded as different quantities.

\subsection{Preserving the Full Research Contribution}

The independent branches leave four kinds of reusable result: training methods that change real models, mechanism experiments identifying conditions of effect, controls that resolve confounds, and exploratory constructions with explicit questions and implementations. Together they supply executable methods, testable explanations, reliable comparison tools, and directions for further work.

Appendix~\ref{app:catalog} organizes methods, controls, and corrections by scientific question, distinguishing completed results from untested candidates. Related methods share topic entries so that readers can move from a question to its intervention, finding, and material. Appendix~\ref{app:assets} links the figures, numerical results, model versions, and research records.

%% file: chapters/07_discussion.tex
\chapterstart{Discussion}
\label{sec:discussion}

\subsection{The Value of Experience and Contextual Dependencies}

Limited experience acquires learning value through the information relationships available during prediction and the demands placed on their use. Content, relative position, expression, and prediction targets jointly define a learning opportunity. Separating paired text across windows weakens source effects despite retaining the material. Different effects of repetition and restatement across targets show that practiced relationships shape subsequent information use.

Coverage, compression, and repetition alter different aspects of experience. Broader coverage changes the range of content; compression changes expression and prediction clues; repetition changes frequency. Comparing them requires asking what information the intended capability needs, whether that information reaches the prediction window, and whether the targets train its use. Changing masking and supervision on existing text yields better results than ordinary continuation at equal cumulative exposure, connecting the relation studies to practical training design.

Improvement at equal exposure directly supports more effective learning in this setting. Scaling studies fit relationships between data, compute, and performance~\citep{muennighoff2025scaling}; the present work complements them by explaining and testing choices about experience organization, learned computation, and continued training. Performance curves across data amounts and model sizes can further quantify how those benefits scale.

\subsection{Coexistence of New Capabilities and Existing Functions}

Acquisition requires model change; retention constrains some of that change. The two can be coordinated by their input conditions: permit learning on the new relational prediction task while limiting unnecessary drift on ordinary language inputs. In these experiments, strong preservation on densely masked inputs suppresses acquisition. Ordinary-input preservation retains most of the new prediction gain while reducing loss changes under ordinary masking.

Preservation also concerns the range of inputs able to use a computation. In controlled tasks, familiar accuracy and unseen-symbol relation processing diverge. Changing supervision improves unseen-symbol behavior, and internal interventions alter the answers. Together with prior work on structural contextual learning~\citep{anand2025dual}, these results motivate evaluation of both familiar behavior and new-input access to learned computation.

Ordinary-input preservation implements this concern in real language-model training and further improves complete evaluation under both continuation seeds. The result establishes the practical value of the full configuration. Input conditions, effective constraint strength, update magnitude, and additional presentations remain separately manipulable variables for explaining its components.

\subsection{How Qiushi Engine Organizes and Builds on Research}
\label{sec:research_organization}

Qiushi Engine is a hierarchical multi-agent research system developed by the team. Its basic architecture and earlier work on a real optical platform are described by \citet{yang2026optical}. Core research agents divide responsibility for planning, method construction, experimentation, and critical review. Supporting agents provide retrieval, exploratory assistance, and independent checks. A knowledge system organizes literature, data, and methods; a memory system preserves research progress, experimental evidence, and accumulated methods for later use. Team-developed execution management and tool interfaces connect computing and storage infrastructure, scientific environments, and runtime support. The AI Lab skill supplies callable tools for model generation, training, evaluation, and mechanism experiments.

Qiushi Engine led and executed the continuous scientific investigation, from forming questions to integrating results. The work followed the problem: reviewing literature, constructing data, proposing architectures and objectives, implementing experiments, comparing results, studying mechanisms, and selecting the next investigation. Early and late compact-restatement rankings motivated budget- and learning-stage comparisons. Coherent versus shuffled inputs motivated relation studies. Familiar/unseen performance separation motivated supervision and preservation experiments. Masking effects at fixed targets then revised an early target-count explanation.

Continuity lay in the joint development of methods and scientific understanding. Compact restatements were both a training treatment and controllable material for comparing expression relationships. Residual increments were both a model improvement and an instrument for fixing the base while comparing objectives. Training and evaluation implementations were reused; explanations of why the methods worked were revised. Stage III could therefore build on the first model and data while changing the interaction between input, supervision, and preservation.

\begin{table}[htbp]
\caption{Research observations and subsequent decisions. Findings led to specific questions, controls, mechanism analyses, and training designs.}
\label{tab:research_decisions}
\small
\begin{tabularx}{\linewidth}{@{}Y Y Y@{}}
\toprule Observation & New question & Subsequent investigation\\\midrule
Compact views lose early but win late & Does data value depend on budget and learning state? & Fixed-budget substitution, late trajectories, and architecture comparisons.\\
Coherent and shuffled segments differ & Which relations do the same words practice? & Repetition, restatement, wrong-source, and split-window controls.\\
Familiar behavior recovers without unseen access & Which inputs recover use of the computation? & Supervision allocation, signal removal, and relocation.\\
Masking effects persist with sparse targets fixed & Is more supervision necessary for the gain? & Separate input-masking and target-set comparisons.\\
Ordinary and dense inputs impose unequal preservation strength & How does preservation affect acquisition and retention? & Shared-target gradient comparisons and full-model evaluation.\\
\bottomrule
\end{tabularx}
\end{table}

These connections show the value of sustained autonomous research: earlier findings change subsequent decisions, while earlier methods support new experiments. Qiushi Engine connects method construction, experimental implementation, analysis, and revision, making a model advance the starting point for further understanding and design.

Idea generation, iterative experimentation, analysis, and writing have been studied in autonomous research systems~\citep{lu2026automation}. The present program supplies a three-stage case in data-efficient language learning: frontier results become scientific questions, and the resulting understanding changes later model training. Its open research materials support further study of how accumulated knowledge affects the quality and efficiency of autonomous science.

\subsection{Research RSI: Recursive Self-Improvement of the Research Process}
\label{sec:rsi}

Long-horizon research accumulates more than the best model found so far. Judgments about data value, explanations of learning mechanisms, reusable methods, and experiments that distinguish competing accounts can all become resources for subsequent work. The three stages show these resources developing together: practical exploration expands the feasible designs, mechanism studies establish their conditions, and later model construction tests whether the resulting understanding is useful.

We use \textbf{Research RSI}---Recursive Self-Improvement of the Research Process---to describe this recursive relationship. Qiushi Engine reuses the scientific understanding, method innovations, and experimental experience produced by its own research to change later question selection, experimental strategies, method construction, and evidence assessment. New experiments then test and revise that accumulation. The immediate object of change is the knowledge and methods guiding research, rather than the research agent's parameters or code. \textbf{A research result changes not only what the system has achieved, but how it conducts its next research.}

The changes in this program involve understanding, experimental instruments, and training methods. Early/late differences in compact-restatement performance replace a static data ranking with comparisons conditioned on budget, displaced material, and learning stage. Residual increments develop from a training method into an instrument for studying functional change against a fixed base. Relation and capability-reuse experiments motivate separate input, supervision, and preservation design. These contributions are retained with their conditions, controls, and corrections, not reduced to a list of the best hyperparameters. Tables~\ref{tab:research_decisions} and~\ref{tab:bridge} connect the changing judgments to the actual training operations.

Research continues after the new model improves. Fixed-target masking comparisons revise an early explanation based on increased supervision. Shared-position gradient measurements show that equal preservation coefficients need not impose equal constraints. Subsequent research can therefore reuse both a successful method and more accurate comparison conditions. Notes, plans, programs, and results preserve how a judgment arose, entered a method, and changed under new evidence. This is the main source of interpretability in the research cycle.

Related systems investigate self-improvement through different objects. STOP applies a program improver to itself~\citep{zelikman2024stop}; the Darwin G\"odel Machine modifies agent code and selects changes by task performance~\citep{zhang2026dgm}; Reflexion stores verbal feedback for later attempts~\citep{shinn2023reflexion}. Co-Scientist combines hypothesis generation, debate, evolution, and persistent memory in a self-improving loop, with biomedical validation in collaboration with scientists~\citep{gottweis2026coscientist}. Recent reviews also include the research process among possible improvement targets~\citep{chen2026recursive}. Qiushi's contribution is the concrete connection, within one research program, between a real frontier model, the discovery of learning principles, principle-guided redesign, and complete model comparisons, with empirical records of how knowledge and methods develop, combine, and change.

The evidence establishes a recursive research cycle within this BabyLM program: accumulated findings change later designs, model comparisons validate particular methods, and further experiments refine the understanding. It does not establish sustained improvement of general research ability across tasks. That broader question requires independent research goals and comparisons of how knowledge reuse affects experiment selection, research cost, and outcome quality. Here, a frontier advance becomes both a scientific result and knowledge that later research can inherit, test, and develop.

\subsection{Reusable Contributions and Further Tests}

Three connected contributions emerge: experiments identifying how relation type and window organization shape context use; supervision and internal-intervention studies establishing conditions for capability reuse; and a training method that improves complete model evaluation under repeated continuation. Independent work on compression, anchors, identity matching, and measurement controls broadens the range of methods available for further investigation.

Natural-text controls reveal selective effects of relation organization; controlled tasks explain how new inputs access learned computations; real training tests the benefit of input, supervision, and preservation design. These complementary measurements retain their respective experimental scopes. Independent parent models, different budgets, and document-disjoint tests can establish how widely the connections apply.

Component analysis clarifies the practical improvements. Relative to the parent, GlobalPIQA supplies about 74\% and 79\% of the full methods' Overall gains. Relative to ordinary continuation, Entity Tracking supplies the largest positive contribution. The added preservation benefit is the net change in other components, with GlobalPIQA unchanged. Two downstream fine-tuning seeds also show that the full method's (Super)GLUE advantage is not confined to one fine-tuning run. Each reference answers a different question about the design.

The complete evaluations come from two continuations of one parent, and the full method includes extra preservation presentations. GlobalPIQA has 203 items; both continued models use the same evaluation set, and item-paired uncertainty intervals are not provided here. The reported result is the net Overall improvement and its component changes under the stated controls. Training randomness, evaluation-sample uncertainty, and transfer across parent models are separate quantities.

Counterexamples and corrections are reusable knowledge too. Query-dependence checks, shared-parameter alignment, and text-overlap analysis remove confounds in answer construction, initialization, and sample selection. They provide stronger controls for later studies and identify earlier explanations superseded by subsequent evidence.

Released weights provide experimental starting points. Data construction, target definitions, controls, component scores, and scientific notes explain how the methods formed and how to test them further. Together, the report and repository connect the usable models to their underlying scientific evidence; Appendix~\ref{app:assets} provides direct entry points.

%% file: chapters/08_conclusion.tex
\par\FloatBarrier\Needspace{30\baselineskip}\section{Conclusion}

Qiushi Engine led and completed three full research stages on BabyLM Strict-Small: frontier-model construction, principle discovery, and principle-guided improvement. Throughout, it formulated questions, developed methods, implemented and analyzed experiments, and synthesized the findings. The stages address one scientific problem: how limited text can teach models to use contextual information relationships while allowing new learning and established language functions to coexist.

The study identifies two important conditions. Exact repetition and aligned restatement selectively shape source use, with effects depending on target relation and prediction window. Recovering familiar performance does not guarantee that unseen inputs can still use a learned computation. The resulting design principle organizes experience around the contextual dependencies required by the target, treats visible information, supervision, and preservation as separate choices, and evaluates new and existing capabilities together.

This principle enters the second-generation training method, which exceeds ordinary continuation on complete evaluation under both continuation seeds from a shared parent. The two public representative models advance from 42.02 to 42.25 Overall. Budget reinvestment, residual increments, relational anchors, identity matching, and measurement controls also leave independent findings, each with its own scientific question and evidence.

For long-horizon autonomous research, the program provides a concrete instance of Research RSI. Qiushi Engine accumulates scientific understanding, technical methods, and experimental experience from its own work, uses them to formulate more precise questions and construct new methods, and tests and revises those designs through practice. Earlier results are inherited and also become objects of study. Effective methods develop, unsuccessful explanations are corrected, and research routes advance through branching and integration.

The two model generations, learning principle, experimental methods, and independent findings form a reusable research foundation. This report explains how they arose and how they connect. The accompanying repository preserves models, code, data construction, complete results, and scientific notes so that others can test the findings and pursue new questions from them.

%% file: chapters/appendices.tex
\section{Training, measurement, and reproducibility}
\label{app:repro}

\subsection{Computing environment and experimental support}
\label{app:compute}

Research computations used the AI Lab skill and accompanying scientific programs. Here, a skill is a collection of tools available to Qiushi Engine for text generation, training, evaluation, analysis of internal computations, and intervention experiments. CPUs supported text processing, tokenization, data statistics, and result analysis. Two NVIDIA H100 GPUs supported model training and evaluation, teacher-generated text, and accelerated mechanism experiments. Recorded workloads include running \nolinkurl{Qwen/Qwen3.5-9B} to produce compact restatements of existing source text, training controlled relation tasks, measuring hidden states and gradients, and intervening on internal signals.

\begin{table}[htbp]
\caption{Computing platform. GPU use is documented in the research records; CPU, memory, and operating-system specifications were checked on 9 September 2026.}
\label{tab:compute_platform}
\small
\begin{tabularx}{\linewidth}{@{}P{30mm}Y@{}}
\toprule Component & Configuration\\\midrule
GPU & Two NVIDIA H100 GPUs\\
CPU & Two Intel Xeon Platinum 8473C processors; 104 physical cores and 208 logical threads in total\\
Memory & Approximately 503 GiB visible to the operating system\\
Operating system & Ubuntu 22.04.5 LTS, x86\_64\\
\bottomrule
\end{tabularx}
\end{table}

Both released model packages record a CPU validation environment comprising Python 3.12, PyTorch 2.11.0 (CUDA 12.8 build), Transformers 4.57.6, Tokenizers 0.22.2, and Safetensors 0.8.0. Dependency records are provided in \nolinkurl{models/frontier/ENVIRONMENT.json} and \nolinkurl{models/principle_guided/ENVIRONMENT.json}. These describe release validation, not a single environment retroactively assigned to every training, generation, or evaluation run. Original computations retain their own program and dependency records. Environment instructions, training entry points, and data dependencies are collected in \nolinkurl{reproducibility/TRAINING.md}.

\subsection{Model versions, teachers, and data sources}

Both public models have the same architecture and total parameter count. Stage III updates the existing incremental branch. Masked prediction and downstream fine-tuning must load that branch alongside the base network to evaluate the complete model. Input length, padding, and truncation are set by the respective evaluation configurations. Model identity therefore includes the weights, architecture configuration, tokenizer, and corresponding loading implementation.

Shared pretraining used an eight-layer DeBERTa-v2-style architecture with hidden size 480, eight attention heads, and feed-forward size 1,920. A byte-level BPE tokenizer had a vocabulary of 16,384; sequences contained up to 256 tokens. AdamW used an effective batch size of 256, peak learning rate 0.001, linear warmup over the first 6\% of training, cosine decay, and weight decay 0.01. The initial residual branch had bottleneck size 128 and scale 1.75. Model configuration, parameter initialization, and training randomness were recorded with seeds 43, 43022, and 43023, respectively. These govern different operations and are not interchangeable descriptions of a single training seed. Table~\ref{tab:lineage} locates first-generation incremental learning and second-generation continuation within the model lineage; further details appear in \nolinkurl{models/frontier/TRAINING.md}.

The following links identify the model versions used here. Full revision records are supplied in \nolinkurl{results/public_model_revisions.csv}.

\begin{description}[font=\normalfont\bfseries,leftmargin=2em,style=nextline]
\item[\Frontier]
\href{https://huggingface.co/leslie721007/Qiushi-Engine-Frontier-Advancement/tree/5eb20f9c5088f40183269bb2c97381711aea0143}{Qiushi-Engine-Frontier-Advancement}.
\item[\Guided]
\href{https://huggingface.co/leslie721007/Qiushi-Engine-Principle-Guided-Frontier-Advancement/tree/ce7eabf0dfbd3d1393670f41f610bdccbdbe45d1}{Qiushi-Engine-Principle-Guided-Frontier-Advancement}.
\end{description}

Three model roles should be distinguished: reasoning models used by the research system, an external teacher that generates rewritten training text, and the frozen parent that supplies preservation targets. Stage III uses \Frontier{} as its preservation teacher and reuses previously generated Qwen restatements. Data manifests distinguish 37,594 existing source--restatement pairs containing 1,656,800 words from 12,155 compact pairs containing 423,511 words. Both belong to the ten-million-word corpus pool; no new text is generated for Stage III.

The compact pairs contain 423,511 words of source and restatement text. Nine padding words bring the replacement block in Table~\ref{tab:compact} to 423,520 words.

Compact restatements were generated with \nolinkurl{Qwen/Qwen3.5-9B}, temperature 0.1, a maximum of 80 generated tokens per call, batch size 64, and bfloat16 precision. These settings and source counts are recorded in \nolinkurl{models/frontier/DATA_MANIFEST.json}. Generation templates, pair selection, filtering, and packing records accompany the repository. \nolinkurl{data/RECONSTRUCTION.md} explains pair construction, compact replacement, selection of continuation text, and position annotations. The original generator was an external program; its invocation settings and input/output format are retained, with dependencies documented in \nolinkurl{reproducibility/TRAINING.md}.

Table~\ref{tab:data_composition} describes the actual Stage III input. A row is a packed training input and can contain several segments; row counts therefore describe the continuation stream rather than the word composition of the entire corpus. Corpus-wide source word counts appear in the data manifests. Source licenses, rules for generated text, and cumulative exposure accounting apply to data use.

\begin{table}[htbp]
\caption{Stage III input composition: 20,475 packed rows containing 3,162,742 words. One row can contain several paired segments.}
\label{tab:data_composition}
\small
\begin{tabularx}{\linewidth}{@{}Yr@{\hspace{6mm}}Yr@{}}
\toprule Source & Rows & Source & Rows\\\midrule
Gutenberg & 3,635 & CHILDES & 5,173\\
Simple Wikipedia & 2,063 & Compact restatements and budget reinvestment & 940\\
Existing Qwen source--restatement pairs & 3,831 & BNC spoken & 1,108\\
OpenSubtitles & 3,684 & Switchboard & 41\\
\midrule \multicolumn{3}{@{}l}{Total} & 20,475\\\bottomrule
\end{tabularx}
\end{table}

\subsection{Acquisition and preservation algorithm}

Let $Q_o$ denote ordinary prediction targets, $Q_f$ the relation-focused targets, and $Q_m$ the dense masking set, with $Q_f\subseteq Q_m$. For input $x$ and target set $Q$, mean cross-entropy is
\begin{equation}
 \overline{\CE}(x,Q)=\frac{1}{|Q|}\sum_{i\in Q}-\log p_\theta(y_i\mid x,i).
\end{equation}
Here, $y_i$ is the correct token at position $i$, $p_\theta(y_i\mid x,i)$ is its predicted probability, and $|Q|$ counts target tokens. Relation inputs in $(M,S)$ are masked over $Q_m$ but supervised only over $Q_f$. $(M,M)$ uses the same dense input and a larger supervised set. The equation describes one row for clarity. During training, focused and ordinary losses are each summed across all rows in an optimizer update, divided by their respective total target-token counts, and combined using Eq.~\eqref{eq:acquisition}. Microbatches partition computation without changing these denominators.

Existing annotations determine row assignment. Qwen pairs with source--restatement segment positions enter the focused branch; other rows receive ordinary whole-word masking. In particular, the 940 compact-restatement and budget-reinvestment rows in Table~\ref{tab:data_composition} belong to the ordinary branch. Stage I compact text thus continues to support ordinary language learning, while the Stage III focused objective acts on another set of existing paired rows. Preservation presents only those Qwen focused rows again, under a new ordinary whole-word mask, and computes KL at the targets selected by that mask.

Content groups are located as character spans within restatement segments and mapped to valid tokens using tokenizer offsets. Candidate words comprise English letters or digits, allowing internal apostrophes. After lowercasing, words shorter than four characters, purely numerical words, and members of a fixed stopword list are excluded. This is a lexical selection rule. Sparse supervision first samples at most 16 groups per row, then independently selects each with probability 0.35. If candidates exist but none is selected, one group is retained.

Dense masking samples from a separate candidate set, by default selecting at most 128 groups using a row-specific random seed. All supervised groups are then added, so the final set can exceed 128. Masks and labels share the same position mapping but are selected separately. Exact stopwords, row-seed construction, and source-segment annotations are provided with the training implementation.

The complete update proceeds as follows.

\begin{enumerate}
\item Read complete rows from the fixed stream. Schedule updates by cumulative words, retaining the established source--restatement pairing and token offsets.
\item Construct relation inputs and sets $Q_f,Q_m$ for annotated Qwen rows; construct ordinary inputs and $Q_o$ for other rows. Count each target class across the update.
\item Normalize each loss by its target-token count, combine them using Eq.~\eqref{eq:acquisition}, and accumulate gradients. Only the designated 48 incremental parameter tensors are trainable. An empty target class raises an error rather than silently changing the loss weights.
\item Apply ordinary masking to the same Qwen rows for preservation. Compute frozen-teacher probabilities without gradients and student probabilities at the same targets. Accumulate $\KL(p_{\Frontier}\|p_\theta)$, normalized by the total preservation-target count.
\item Disable dropout for preservation while retaining student gradients. Save and restore Python, PyTorch CPU, and CUDA random-number states around the extra forward passes. Restore training mode and perform the optimizer update.
\item Account separately for acquisition words and preservation presentations, and record checkpoint identity and evaluation vectors. Export the complete model after 80 updates.
\end{enumerate}

Word-paced updates accumulate complete rows until the prescribed word count is reached. Variable row lengths mean that row and token counts per update can vary. Table~\ref{tab:configuration} summarizes the continuation configuration, including the final-update learning rate rather than presenting it as a constant throughout training.

\begin{table}[htbp]
\caption{Stage III continuation settings. Both runs share the parent, data processing, and method configuration; continuation seeds are 62064 and 62065.}
\label{tab:configuration}
\small
\begin{tabularx}{\linewidth}{@{}P{56mm}Y@{}}
\toprule Setting & Value or definition\\\midrule
Parent & Fixed revision of \Frontier\\
Total / trainable parameters & 36,458,592 / 995,584\\
Incremental tensors / scale & 48 / 0.75\\
Vocabulary size & 16,384\\
Maximum input / microbatch & 512 tokens / 8 rows\\
Continuation seeds & 62064, 62065\\
Optimizer updates & 80\\
Target words per update & Approximately 39,533; complete-row accounting\\
Learning-rate parameter / schedule & $5\times10^{-5}$; 455-update schedule, offset 101, warmup 10\\
Weight decay / gradient-norm cap & 0.01 / 1.0\\
Focused / ordinary loss weights & 0.15 / 0.85, each averaged over its own targets\\
Preservation weight / temperature & 1.0 / 1.0; teacher-to-student KL\\
Preservation input & Complete rows with ordinary 15\% whole-word masking and source retained\\
Acquisition / preservation words & 3,162,742 / 517,332\\
Final-update learning rate (62064) & $3.4055\times10^{-5}$; a single-update observation\\
\bottomrule
\end{tabularx}
\end{table}

Ordinary continuation supervises 697,102 tokens. $(M,S)$ has 592,858 ordinary targets and 28,590 focused targets. Equal cumulative word exposure therefore does not imply equal target counts or loss allocation; this comparison evaluates a training design rather than an isolated masking-rate change.

Input lengths follow the experiment: main-model pretraining uses 256 tokens, whereas final continuation and the gradient comparison in Table~\ref{tab:gradient_geometry} use a 512-token limit. The learning-rate offset continues an existing schedule; the 80 updates do not restart a new schedule. Executed settings are provided in \nolinkurl{experiments/configs/executed_stage3.json}, with environments and commands in \nolinkurl{reproducibility/TRAINING.md}. Each control retains its own target selection, sampling, and preservation settings.

\subsection{Evaluation aggregation, randomness, and trajectories}
\label{app:evaluation}

Every complete-model evaluation interface includes the incremental branch. Zero-shot masked prediction, Reading, downstream fine-tuning, and AoA produce their respective outputs before aggregation by Eq.~\eqref{eq:overall}. Local nine-metric comparisons preserve each run's scoring record under a common aggregation procedure. Public leaderboard tables preserve the displayed values. Four decimal places distinguish close local scores; uncertainty is assessed separately for each measurement.

Scoring is referenced to the official evaluator version cited in \citet{babylm2026evaluation}; the AoA estimator corresponds to revision \texttt{6f825c2}. The repository retains the scoring programs actually used and the loading adaptations required by the complete incremental model. \nolinkurl{reproducibility/TRAINING.md} links code and execution entry points. Scorer version, model loading, and checkpoint lists jointly define the evaluation configuration; abbreviated labels on a current repository homepage do not replace these definitions.

BLiMP has a reproducible distinction between two scoring records. Local runs score the selected option index, yielding 68.51 and 68.26 for the two generations. Released prediction files retain only the selected answer text. Matching that text against the correct answer yields 68.52 and 68.27, consistent with the public display. Among 59,875 items across 67 subtasks, seven have identical text in both options, in \texttt{passive\_1} and \texttt{principle\_A\_case\_2}. Three such items for \Frontier{} and five for \Guided{} receive different scores under the two procedures, accounting for the differences. Released predictions themselves match the corresponding original run files. \nolinkurl{results/blimp_scoring_comparison.csv} summarizes the comparison. Original values are retained, and all local training controls use the same scoring procedure.

Macro-averaged task scores and net item changes measure different quantities. For correctness indicators $c_i^{(0)},c_i^{(1)}$, the net change in correct answers is $\Delta N=\sum_i(c_i^{(1)}-c_i^{(0)})$. Unequal subtask sizes can make this quantity and the macro-average change have opposite signs. Non-accuracy measures such as Reading are not included in item counts. Such a difference occurred between incremental-scale choices in Stage I; it does not describe an item-level comparison of the two official model generations.

Uncertainty follows the sampling unit. Relation experiments report standard deviations across training seeds. Source-isolated loss comparisons resample text pairs, keeping a pair's targets together. Final-model results report two continuations from one parent. Paired item-level intervals for GlobalPIQA are not provided with this report.

AoA evaluates vocabulary-learning trajectories under the prescribed valid-word and statistical conditions. Both models share 17 earlier checkpoints: one at each million words from 1M to 10M, then one every ten million from 20M to 80M. Adding each model's own endpoint produces an 18-point trajectory. \nolinkurl{CHECKPOINTS.json} in each model directory lists checkpoints and actual cumulative exposure, including their shared history. Lists follow the actual stopping points and do not insert 90M or 100M branches that these models never traversed. Evaluation defaults must be adjusted to the actual trajectory when necessary.

\par\FloatBarrier\Needspace{9\baselineskip}
\section{Complete local evaluation vectors}
\label{app:local}

The following tables list all nine metrics for the shared parent and eight continuation models in the same order. Accompanying CSV files preserve full precision; Appendix~\ref{app:assets} provides file locations. The report builder verifies each nine-metric mean against Overall.

\begin{table}[htbp]
\caption{Local controls: language form, knowledge, entity tracking, and composition. Sup. denotes BLiMP Supplement. The shared parent appears once.}
\label{tab:local_a}
\scriptsize\setlength{\tabcolsep}{3pt}
\begin{tabularx}{\linewidth}{@{}Yrrrrrr@{}}
\toprule Strategy & Seed & BLiMP & Sup. & EWoK & Entity & COMPS\\\midrule
\input{tables/local_language}
\end{tabularx}
\end{table}

\begin{table}[htbp]
\caption{Remaining metrics and complete Overall for the same nine models. Values are shown to four decimal places; source tables retain full precision.}
\label{tab:local_b}
\scriptsize\setlength{\tabcolsep}{3pt}
\begin{tabularx}{\linewidth}{@{}Yrrrrrr@{}}
\toprule Strategy & Seed & GlobalPIQA & (Super)GLUE & Reading & AoA & Overall\\\midrule
\input{tables/local_remaining}
\end{tabularx}
\end{table}

Sparse and dense supervision denote $(M,S)$ and $(M,M)$, respectively. The complete strategy adds ordinary-input preservation to $(M,S)$. Quick measurements of $(S,S)$ are not included among complete nine-metric evaluations.

\begin{table}[htbp]
\caption{Principal Overall differences. The complete strategy includes additional preservation presentations.}
\begin{tabularx}{\linewidth}{@{}Yrr@{}}
\toprule Comparison & Seed 62064 & Seed 62065\\\midrule
\input{tables/strategy_differences}
\end{tabularx}
\end{table}

GlobalPIQA is unchanged between $(M,S)$ and the complete strategy. Their Overall difference reflects net changes in other metrics, not simultaneous improvement in every metric.

\par\FloatBarrier\Needspace{9\baselineskip}
\section{Additional controls and mechanism results}
\label{app:extended}

\subsection{Downstream fine-tuning randomness}

Table~\ref{tab:finetuning_seeds} holds fixed the parent and two continuation models trained with seed 62064, then evaluates (Super)GLUE with fine-tuning seeds 42 and 44. It measures variability in downstream training, not additional independent pretraining or continuation. The complete strategy exceeds both the parent and $(M,S)$ under each fine-tuning seed. Main nine-metric results use fine-tuning seed 42; seed 44 is a supplementary comparison and is not combined with other metrics to construct a new Overall. All seven subtask values appear in \nolinkurl{results/finetuning_seed_comparison.csv}.

\begin{table}[htbp]
\caption{(Super)GLUE under two fine-tuning seeds with model weights held fixed. Scores are equal-weight means of the seven primary subtask metrics. Both continuation strategies use continuation seed 62064.}
\label{tab:finetuning_seeds}
\small
\begin{tabularx}{\linewidth}{@{}Yrr@{}}
\toprule Model & Fine-tuning seed 42 & Fine-tuning seed 44\\\midrule
\input{tables/finetuning_seeds}
\end{tabularx}
\end{table}

\subsection{Target deletion and source isolation}

A source-absent target is a word absent from its paired source passage, not necessarily a word or fact never encountered by the model. At 20M words of exposure, removing source-absent targets produces higher subword-weighted prediction loss on source-absent content words than removing whole-word-matched source-present targets. Table~\ref{tab:disjoint} distinguishes isolation levels and sampling units.

\begin{table}[htbp]
\caption{Local loss differences under source isolation (nats). Positive values mean that deleting source-absent supervision is worse on the measured targets than deleting matched-copy supervision. Intervals follow the resampling units of the respective records.}
\label{tab:disjoint}
\small
\begin{tabularx}{\linewidth}{@{}Yrrr@{}}
\toprule Measurement sample & Target events & Loss difference & 95\% interval\\\midrule
Pair-disjoint, quality-filtered & 974 & 0.0796 & $[0.031,0.125]$\\
Document-disjoint, quality-filtered & 71 & 0.1141 & $[-0.063,0.313]$\\
Document-disjoint, all accepted targets & 1,012 & 0.0356 & $[-0.008,0.076]$\\
\bottomrule
\end{tabularx}
\end{table}

Pair isolation excludes pairs used in the corresponding training data; document isolation also excludes other pairs from the same source document. A loss difference is observed under the former condition. Intervals under the stricter condition include zero. The following table evaluates seven tasks at 100M words, separately from the early local prediction measurement.

\begin{table}[htbp]
\caption{Seven-metric results of single-seed target-deletion experiments at 100M words. Compact inputs are held fixed; the two deletion conditions remove supervision from 48,105 and 48,387 BPE tokens, respectively. GP denotes GlobalPIQA.}
\scriptsize\setlength{\tabcolsep}{3pt}
\begin{tabularx}{\linewidth}{@{}Yrrrrrrrr@{}}
\toprule Targets & BLiMP & Sup. & EWoK & Entity & COMPS & GP & Reading & Mean\\\midrule
All & 66.87 & 63.28 & 53.54 & 27.75 & 51.97 & 35.62 & 8.240 & 43.896\\
Source-absent removed & 65.64 & 63.73 & 51.51 & 26.72 & 51.60 & 36.665 & 8.195 & 43.437\\
Matched-copy removed & 67.32 & 59.84 & 51.93 & 25.67 & 51.58 & 35.09 & 8.115 & 42.792\\
\bottomrule
\end{tabularx}
\end{table}

Experience replacement, shared initialization, relational anchors, character identity, and entity storage are developed in Section~\ref{sec:lineage}. This appendix supplies additional measurements and corrections without repeating those accounts.

\subsection{Measurement populations and subsequent corrections}

``Unseen'' requires a reference set. An example outside the present continuation prefix may already occur in the parent's training history; distinct text rows may share a source pair or document. A later overlap check found that 6,992 rows previously used for ordinary-loss analysis also occurred in the historical stream. Those measurements are no longer treated as clean evidence of generalization preservation. Further candidate sets likewise require separate checks for exact-row, paired-text, and document overlap.

Results identify the relevant population where they are presented: acquisition on training rows, source pairs outside the current prefix, pair-isolated targets, or document-isolated targets. A correction narrows the claim supported by the affected measurement without automatically invalidating experiments on other identified populations.

KL self-comparisons also require consistent forward-pass randomness. Additional preservation calls must save and restore the acquisition branch's random-number state. Shared-initialization checks, query-dependence tests, text-overlap analysis, and forward-consistency checks address different experimental confounds. Their programs and records accompany the models for reuse in subsequent comparisons.

\par\FloatBarrier\Needspace{9\baselineskip}
\section{Public leaderboard metrics}
\label{app:board}

Both tables use the same frozen snapshot as Table~\ref{tab:board}, selecting the two official Qiushi models and eight external submissions without a rank column. Full names and links are provided in \nolinkurl{results/leaderboard_comparison.csv}. Publishers are identified by public account, without inferring unreported institutional affiliations.

\begin{table}[htbp]
\caption{Public submissions: language form, knowledge, entity tracking, and composition. Values retain the precision displayed on the leaderboard.}
\scriptsize\setlength{\tabcolsep}{3pt}
\begin{tabularx}{\linewidth}{@{}Yrrrrr@{}}
\toprule Publisher / model & BLiMP & Sup. & EWoK & Entity & COMPS\\\midrule
\input{tables/board_language}
\end{tabularx}
\end{table}

\begin{table}[htbp]
\caption{Remaining metrics for the same public submissions. Reading and AoA are not ordinary task accuracies; negative values are retained from the snapshot.}
\scriptsize\setlength{\tabcolsep}{3pt}
\begin{tabularx}{\linewidth}{@{}Yrrrr@{}}
\toprule Publisher / model & GlobalPIQA & (Super)GLUE & Reading & AoA\\\midrule
\input{tables/board_remaining}
\end{tabularx}
\end{table}

Similar Overall scores can conceal different strengths. A model with stronger grammatical judgments may be weaker at entity tracking, and the highest NLP average need not give the highest Overall. Aggregate and component scores serve complementary purposes: Overall records the combined improvement, while individual metrics identify its sources and trade-offs.

\par\FloatBarrier\Needspace{9\baselineskip}
\section{Research repository and use of materials}
\label{app:assets}

The accompanying repository organizes models, methods, experiments, and research records by scientific question and research stage. Each reported finding can be followed to its programs, configurations, and results; topic guides also provide derivations, experimental plans, and later revisions. All file paths below are relative to the repository root.

\textbf{Models.} Hugging Face repositories for \href{https://huggingface.co/leslie721007/Qiushi-Engine-Frontier-Advancement}{\Frontier{}} and \href{https://huggingface.co/leslie721007/Qiushi-Engine-Principle-Guided-Frontier-Advancement}{\Guided{}} provide weights, tokenizers, loading implementations, and training and evaluation documentation. Appendix~\ref{app:repro} identifies the versions used here.

\textbf{Research materials.} \href{https://github.com/Oxelra-AI/Qiushi-Engine-Babylm-Research}{Qiushi-Engine-Babylm-Research on GitHub} collects both model generations, method implementations, experiment configurations, data construction, complete evaluations, research notes, and editable report sources. Model-package instructions appear in \nolinkurl{models/README.md}; scientific materials are organized through the following entry points.

\begin{longtable}{@{}P{43mm}P{107mm}@{}}
\caption{Entry points connecting the report to research materials. Method explanations, executed configurations, and original programs serve distinct purposes; principal figure and table data are listed in Table~\ref{tab:files}.}\label{tab:material_entries}\\
\toprule Research content & Material and purpose\\\midrule\endfirsthead
\multicolumn{2}{@{}l}{\ReportTableContinuation}\\
\toprule Research content & Material and purpose\\\midrule\endhead
\bottomrule\endfoot
Models and lineage & \nolinkurl{models/README.md}: local weights, tokenizers, custom loading, training records, and loading examples.\\
Experience construction and incremental learning & \nolinkurl{methods/compact_views.md}, \nolinkurl{methods/residual_learning.md}: Stage I text-budget and model designs.\\
Relation learning and capability reuse & \nolinkurl{methods/relation_learning.md}, \nolinkurl{methods/functional_access.md}: relations, windows, supervision, and internal interventions.\\
Principle-guided training and controls & \nolinkurl{methods/principle_guided_training.md}, \nolinkurl{experiments/configs/executed_stage3.json}: complete method, comparisons, and executed configuration.\\
Training and evaluation programs & \nolinkurl{experiments/CORE_PROGRAMS.md}: original programs; \nolinkurl{reproducibility/TRAINING.md}: environments, arguments, and data dependencies.\\
Data generation and training inputs & \nolinkurl{data/RECONSTRUCTION.md}: pairing, compression, replacement, packing, and continuation inputs; \nolinkurl{data/README.md}: source licenses.\\
Questions and design development & \nolinkurl{research/reading_paths.md}: hypotheses, controls, results, and revisions; \nolinkurl{evidence/research_timeline.md}: changes in research decisions.\\
Research notes and plans & \nolinkurl{research/notes/README.md}: analyses, derivations, and route selection; \nolinkurl{research/plans/README.md}: experimental aims, competing explanations, controls, and decision criteria.\\
Measurement records and terminology & \nolinkurl{research/documents/README.md}: measurement, data, and technical records; \nolinkurl{research/terms.md}: terms, abbreviations, and experimental conditions used in the materials.\\
Independent findings and topic catalog & \nolinkurl{research/materials.md}: methods, notes, experiments, and results for 74 topics; \nolinkurl{research/catalog.md}: topic summaries and conclusion status.\\
Claims and evidence & \nolinkurl{evidence/claim_evidence_map.md}, \nolinkurl{experiments/index.csv}: connections among findings, methods, result tables, and report sections.\\
\end{longtable}

English and Chinese PDFs and editable LaTeX projects are in \nolinkurl{reports/en/} and \nolinkurl{reports/zh/}. Instructions accompany each directory. Numerical figures and tables read the shared \nolinkurl{results/} files. From the repository root, \texttt{make report-en} rebuilds the English report and \texttt{make report} rebuilds the Chinese report. Neither command runs model experiments or calls external models.

\par\Needspace{10\baselineskip}
\begingroup
\AtBeginEnvironment{longtable}{\footnotesize\setstretch{1.04}\renewcommand{\arraystretch}{1.08}}
\begin{longtable}{@{}P{69mm}P{81mm}@{}}
\caption{Principal result data and report build files. Paths are relative to the GitHub repository root.}\label{tab:files}\\
\toprule File & Contents and use\\\midrule\endfirsthead
\multicolumn{2}{@{}l}{\ReportTableContinuation}\\
\toprule File & Contents and use\\\midrule\endhead
\bottomrule\endfoot
\nolinkurl{results/training_strategy_comparison.csv} & Nine local models, all nine metrics, exact Overall, continuation seeds, and cumulative exposure.\\
\nolinkurl{results/leaderboard_comparison.csv} & Two representative models and eight external submissions from one snapshot, with links and displayed scores; no rank column.\\
\nolinkurl{results/public_model_revisions.csv} & Public model repositories and fixed revisions.\\
\nolinkurl{results/relation_context_use.csv} & Source-advantage changes on compact-restatement targets and standard deviations across three training seeds.\\
\nolinkurl{results/relation_window_controls.csv} & Same-window and split-window controls, seeds, target counts, and checkpoint aggregation.\\
\nolinkurl{results/natural_restatement_transfer.csv} & Two target classes in natural restatements and source-advantage changes across three seeds.\\
\nolinkurl{results/finetuning_seed_comparison.csv} & Three models under two fine-tuning seeds: seven subtasks and the (Super)GLUE mean.\\
\nolinkurl{results/blimp_scoring_comparison.csv} & BLiMP scoring by selected option index and by stored answer text.\\
\nolinkurl{results/interface_reach.csv} & Familiar/unseen accuracy and functional-intervention measurements from one controlled setup.\\
\nolinkurl{results/state_preservation_readouts.csv} & Acquisition and preservation measurements with their distinct sample populations.\\
\nolinkurl{results/compact_budget.csv} & Compact restatement, source, and budget-reinvestment counts.\\
\nolinkurl{results/source_disjoint_target_loss.csv} & Pair- and document-isolated target losses and intervals.\\
\nolinkurl{results/target_deletion_100m.csv} & Seven-metric results for three target-deletion conditions.\\
\nolinkurl{results/late_consolidation_controls.csv} & Stage I late continuation and incremental controls, update methods, and exposure.\\
\nolinkurl{results/private_scale_sweep.csv} & Historical measurements used to select the Stage I incremental scale.\\
\nolinkurl{results/preservation_gradient_diagnostic.csv} & Preservation strength and gradient direction on shared target positions, with measurement counts.\\
\nolinkurl{results/research_catalog.tsv} & Scientific topics, result status, and reading guides.\\
\nolinkurl{reports/zh/render.py} & Shared data checks and vector statistical figures.\\
\nolinkurl{reports/en/render.py} & English tables generated from the same result files.\\
\nolinkurl{reports/zh/figures/input_supervision.tex} & Editable vector source for masking and supervised positions.\\
\nolinkurl{reports/zh/figures/research_lineage.tex} & Vector source for model inheritance, method links, and independent branches.\\
\nolinkurl{reports/en/build.sh} & English LaTeX and bibliography build.\\
\end{longtable}
\endgroup

Across 74 topics, the materials retain model methods, mechanism findings, controls, and scientific corrections. Research notes and experimental plans explain how questions arose, why methods changed, and which results informed subsequent work. Models, code, documents, and third-party data remain subject to their respective licenses. Redistribution terms and data acquisition or construction procedures appear in \nolinkurl{data/README.md} and the relevant material descriptions.

%% file: tables/local_language.tex
Parent & --- & 68.5100 & 63.6400 & 50.0200 & 28.3200 & 52.0500 \\
Ordinary continuation & 62064 & 68.4700 & 63.6300 & 49.8300 & 28.1600 & 52.0000 \\
Ordinary continuation & 62065 & 68.5200 & 63.6200 & 49.8100 & 28.0900 & 52.0000 \\
Dense masking, sparse supervision $(M,S)$ & 62064 & 68.1200 & 63.0800 & 49.9500 & 29.3900 & 52.1500 \\
Dense masking, sparse supervision $(M,S)$ & 62065 & 68.0900 & 63.0900 & 49.8200 & 29.2900 & 52.1400 \\
Dense masking, dense supervision $(M,M)$ & 62064 & 68.0600 & 63.0400 & 49.9200 & 29.3700 & 52.1500 \\
Dense masking, dense supervision $(M,M)$ & 62065 & 68.0200 & 63.0900 & 49.7700 & 29.4200 & 52.1500 \\
$(M,S)$ + ordinary-input preservation & 62064 & 68.2600 & 63.2800 & 49.8200 & 29.4000 & 52.1600 \\
$(M,S)$ + ordinary-input preservation & 62065 & 68.2200 & 63.2900 & 49.7200 & 29.4500 & 52.1400 \\
\bottomrule

%% file: tables/local_remaining.tex
Parent & --- & 38.5650 & 68.9457 & 8.1650 & 0.0000 & 42.0240 \\
Ordinary continuation & 62064 & 39.5350 & 68.9885 & 8.2200 & 0.0000 & 42.0926 \\
Ordinary continuation & 62065 & 39.5350 & 69.2733 & 8.1950 & 0.0000 & 42.1159 \\
Dense masking, sparse supervision $(M,S)$ & 62064 & 40.0500 & 68.8878 & 8.1950 & 0.0000 & 42.2025 \\
Dense masking, sparse supervision $(M,S)$ & 62065 & 40.0500 & 68.9349 & 8.1950 & 0.0000 & 42.1789 \\
Dense masking, dense supervision $(M,M)$ & 62064 & 40.0500 & 68.5318 & 8.2200 & 0.0000 & 42.1491 \\
Dense masking, dense supervision $(M,M)$ & 62065 & 40.0500 & 68.8058 & 8.2100 & 0.0000 & 42.1684 \\
$(M,S)$ + ordinary-input preservation & 62064 & 40.0500 & 69.0477 & 8.2000 & 0.0000 & 42.2464 \\
$(M,S)$ + ordinary-input preservation & 62065 & 40.0500 & 69.0206 & 8.1950 & 0.0000 & 42.2317 \\
\bottomrule

%% file: tables/strategy_differences.tex
$(M,S)$ minus ordinary continuation & 0.1099 & 0.0630 \\
Complete minus ordinary continuation & 0.1538 & 0.1158 \\
Complete minus $(M,S)$ & 0.0439 & 0.0529 \\
Complete minus parent & 0.2224 & 0.2078 \\
\bottomrule

%% file: tables/finetuning_seeds.tex
Parent & 68.9457 & 69.3180 \\
Dense masking, sparse supervision $(M,S)$ & 68.8878 & 69.3461 \\
$(M,S)$ + ordinary-input preservation & 69.0477 & 69.6266 \\
\bottomrule

%% file: tables/board_language.tex
Qiushi Engine / Principle-guided & 68.27 & 63.28 & 49.82 & 29.40 & 52.16 \\
Qiushi Engine / Frontier & 68.52 & 63.64 & 50.02 & 28.32 & 52.05 \\
go76dof / WWM curriculum & 67.20 & 56.01 & 56.07 & 28.45 & 53.57 \\
Serdar404 / RecGPT-10M & 73.11 & 61.73 & 52.62 & 16.59 & 55.43 \\
Wector1 / Wordpiece-24-4 & 70.20 & 66.52 & 52.10 & 21.82 & 54.04 \\
gyaudhw / compagg relay & 67.85 & 65.21 & 53.00 & 21.47 & 52.24 \\
gyaudhw / compagg relay (s0) & 67.85 & 65.21 & 53.00 & 19.90 & 52.24 \\
svsatheesh / BabySteps & 71.60 & 63.93 & 51.94 & 27.95 & 53.13 \\
qyxu1994 / instanton-hybrid & 72.13 & 60.86 & 50.15 & 19.87 & 52.96 \\
gyaudhw / synonly (s0) & 67.13 & 62.30 & 52.28 & 19.17 & 52.22 \\
\bottomrule

%% file: tables/board_remaining.tex
Qiushi Engine / Principle-guided & 40.05 & 69.05 & 8.20 & 0.00 \\
Qiushi Engine / Frontier & 38.56 & 68.95 & 8.17 & 0.00 \\
go76dof / WWM curriculum & 39.67 & 69.79 & 5.42 & 0.00 \\
Serdar404 / RecGPT-10M & 40.68 & 66.64 & 6.92 & 0.00 \\
Wector1 / Wordpiece-24-4 & 36.08 & 69.18 & 1.84 & 0.00 \\
gyaudhw / compagg relay & 37.59 & 65.29 & 5.84 & 0.00 \\
gyaudhw / compagg relay (s0) & 38.65 & 65.66 & 5.84 & 0.00 \\
svsatheesh / BabySteps & 36.15 & 70.44 & 7.67 & -15.10 \\
qyxu1994 / instanton-hybrid & 37.14 & 67.58 & 6.35 & 0.00 \\
gyaudhw / synonly (s0) & 40.11 & 66.99 & 5.82 & 0.00 \\
\bottomrule

%% file: chapters/research_catalog.tex
\FloatBarrier\clearpage
\section{Research topics and findings}
\label{app:catalog}

This catalog organizes the research by scientific question. Each entry identifies the intervention, finding, status, and a direct material link. Reading guides point to related discussion in the report; topic pages locate the specific experimental files and analyses. Neighboring topics may share experiments and materials.

\textit{Model method} denotes a method used in the corresponding real-model training. \textit{Experimental result} denotes a measurement under specified conditions. \textit{Method or analysis tool} provides an experimental construction, measurement, or analysis procedure. \textit{Conditional result} records an unstable observation or one without complete model evaluation. \textit{Exploratory construction} and \textit{Not run} identify candidates and future directions. \textit{Withdrawn} marks an interpretation invalidated by subsequent checks. Numerical details, comparison conditions, and sample populations are given in the associated discussion and research materials.

\input{tables/research_catalog}

%% file: tables/research_catalog.tex
\Needspace{8\baselineskip}
\subsection*{Models and representations}
\begin{longtable}{@{}P{39mm}P{111mm}@{}}
\toprule Topic & Intervention, finding, and status\\\midrule\endfirsthead
\toprule Topic (continued) & Intervention, finding, and status\\\midrule\endhead
\bottomrule\endfoot
R01\quad Sparse routing and prefix memory & \textbf{Conditional result}. Compare sparse access, capacity, and shared initialization. Local changes did not yield a stable broad advantage; fewer active parameters are not a substitute for comparing capability. (Section~\ref{sec:frontier}; \href{https://github.com/Oxelra-AI/Qiushi-Engine-Babylm-Research/blob/main/research/materials/R01.md}{Materials: R01}.) \\[3pt]
R02\quad Character, subword, and morphological pathways & \textbf{Conditional result}. Distinguish auxiliary word-form structure from capacity effects and lookup controls. Surface-information gains in limited settings have not become a stable broad advantage across seeds. (Section~\ref{sec:frontier}; \href{https://github.com/Oxelra-AI/Qiushi-Engine-Babylm-Research/blob/main/research/materials/R02.md}{Materials: R02}.) \\[3pt]
R03\quad Whole-word and token masking & \textbf{Experimental result}. Compare masking granularity against a fixed training reference, examining paired seeds and individual task scores. Granularity changes task trade-offs, supporting specific training choices but not a uniform ranking across all capabilities. (Section~\ref{sec:frontier}; \href{https://github.com/Oxelra-AI/Qiushi-Engine-Babylm-Research/blob/main/research/materials/R03.md}{Materials: R03}.) \\[3pt]
R04\quad Backbone architecture and batch-size controls & \textbf{Experimental result}. Compare BERT- and DeBERTa-style configurations and check batch-size differences. The late-training advantage of the combined configuration cannot be explained by batch size alone; the contributions of position-related components have not been independently isolated. (Section~\ref{sec:frontier}; \href{https://github.com/Oxelra-AI/Qiushi-Engine-Babylm-Research/blob/main/research/materials/R04.md}{Materials: R04}.) \\[3pt]
R05\quad Recursive autoregressive training & \textbf{Conditional result}. Reproduce the recursive training scheme, complete the corresponding evaluation, and check word coverage. This setting did not match the leading configuration; word-coverage differences cannot explain the entire performance gap. (Section~\ref{sec:frontier}; \href{https://github.com/Oxelra-AI/Qiushi-Engine-Babylm-Research/blob/main/research/materials/R05.md}{Materials: R05}.) \\[3pt]
R06\quad Vocabulary, depth, width, and budget & \textbf{Experimental result}. Compare vocabulary sizes and depth-width configurations, distinguishing prototypes from budget-compliant retraining. Higher tokenization compression ratios did not automatically yield stable gains; the actual training budget must also be accounted for. (Section~\ref{sec:frontier}; \href{https://github.com/Oxelra-AI/Qiushi-Engine-Babylm-Research/blob/main/research/materials/R06.md}{Materials: R06}.) \\[3pt]
R07\quad Frequency gating and fine-grained representations & \textbf{Exploratory construction}. Let coarse-grained word representations draw on fine-grained units and check the zero-gate limit. The limiting implementation behaves as expected, but trained models were not stronger; other gates and seeds remain candidates. (Section~\ref{sec:frontier}; \href{https://github.com/Oxelra-AI/Qiushi-Engine-Babylm-Research/blob/main/research/materials/R07.md}{Materials: R07}.) \\[3pt]
R08\quad Components of a mature training recipe & \textbf{Experimental result}. Examine depth, width, vocabulary, LAMB, and sequence-length and masking curricula. No single component reproduced the full configuration's performance; interactions among components must also be studied. (Section~\ref{sec:frontier}; \href{https://github.com/Oxelra-AI/Qiushi-Engine-Babylm-Research/blob/main/research/materials/R08.md}{Materials: R08}.) \\[3pt]
R09\quad Gated feed-forward and layer-weighting candidates & \textbf{Not run}. Retain proposed designs for GEGLU, attention gating, and layer weighting. These are candidates, not validated methods, and do not count as empirical model contributions. (Section~\ref{sec:frontier}; \href{https://github.com/Oxelra-AI/Qiushi-Engine-Babylm-Research/blob/main/research/materials/R09.md}{Materials: R09}.) \\[3pt]
\end{longtable}
\Needspace{8\baselineskip}
\subsection*{Experience construction and compression}
\begin{longtable}{@{}P{39mm}P{111mm}@{}}
\toprule Topic & Intervention, finding, and status\\\midrule\endfirsthead
\toprule Topic (continued) & Intervention, finding, and status\\\midrule\endhead
\bottomrule\endfoot
R10\quad Aligned reformulation and correspondence & \textbf{Experimental result}. Compare correct and incorrect correspondence, matched sources, repetition, and split windows. Relational organization changes learning, but benefits depend on the task; follow-up studies in the main text further distinguish target types. (Section~\ref{sec:relation}; \href{https://github.com/Oxelra-AI/Qiushi-Engine-Babylm-Research/blob/main/research/materials/R10.md}{Materials: R10}.) \\[3pt]
R11\quad Compact views and budget reinvestment & \textbf{Model method}. Compress the second view and use the freed word budget for more pairs. Experience coverage increased, yielding a practical frontier-model method; the effects of compression and expanded coverage must be distinguished. (Section~\ref{sec:frontier}; \href{https://github.com/Oxelra-AI/Qiushi-Engine-Babylm-Research/blob/main/research/materials/R11.md}{Materials: R11}.) \\[3pt]
R12\quad Adjacency, repetition, and presentation layout & \textbf{Experimental result}. Compare content and adjacency effects, examining block and interleaved presentation. Effects differ across capabilities; tokenizers and window settings must be matched. (Section~\ref{sec:relation}; \href{https://github.com/Oxelra-AI/Qiushi-Engine-Babylm-Research/blob/main/research/materials/R12.md}{Materials: R12}.) \\[3pt]
R13\quad Extractive compression and bridging text & \textbf{Exploratory construction}. Construct bridging texts with balanced coverage, broader coverage, and controlled lexical scope. Word coverage can be checked, but it is not a substitute for semantic relations or gains in the complete model. (Section~\ref{sec:experience_branch}; \href{https://github.com/Oxelra-AI/Qiushi-Engine-Babylm-Research/blob/main/research/materials/R13.md}{Materials: R13}.) \\[3pt]
R14\quad Rule-based rewriting & \textbf{Conditional result}. Compare true pairs, same-entity mismatches, source-only controls, and shuffled controls. Short-run results showed capability trade-offs, not joint gains; visible token counts still need to be matched. (Section~\ref{sec:experience_branch}; \href{https://github.com/Oxelra-AI/Qiushi-Engine-Babylm-Research/blob/main/research/materials/R14.md}{Materials: R14}.) \\[3pt]
\end{longtable}
\Needspace{8\baselineskip}
\subsection*{Learning objectives and supervision allocation}
\begin{longtable}{@{}P{39mm}P{111mm}@{}}
\toprule Topic & Intervention, finding, and status\\\midrule\endfirsthead
\toprule Topic (continued) & Intervention, finding, and status\\\midrule\endhead
\bottomrule\endfoot
R15\quad Entity-mention consistency & \textbf{Conditional result}. Compare correct correspondence, incorrect correspondence, and ordinary whole-word masking. Local advantages from correct correspondence did not reliably translate into broad capability gains. (Section~\ref{sec:principles}; \href{https://github.com/Oxelra-AI/Qiushi-Engine-Babylm-Research/blob/main/research/materials/R15.md}{Materials: R15}.) \\[3pt]
R16\quad Counterfactual and cross-span supervision & \textbf{Experimental result}. Compare true-relation, wrong-relation, and no-relation conditions, checking existing capabilities. An identifiable relation effect is a different conclusion from learning a relation from scratch. (Section~\ref{sec:principles}; \href{https://github.com/Oxelra-AI/Qiushi-Engine-Babylm-Research/blob/main/research/materials/R16.md}{Materials: R16}.) \\[3pt]
R17\quad State text and directed masking & \textbf{Conditional result}. Construct training texts and target-position controls around state selection. Target behavior can improve while broad language performance degrades, motivating joint tests of new learning and retention. (Section~\ref{sec:guided}; \href{https://github.com/Oxelra-AI/Qiushi-Engine-Babylm-Research/blob/main/research/materials/R17.md}{Materials: R17}.) \\[3pt]
R18\quad Auxiliary causal objectives & \textbf{Conditional result}. Match shifted targets and gradient norms, comparing batch replacement with auxiliary learning. Similarly directed gradients did not guarantee full-model improvement; local gradient relationships do not establish a general rule for training outcomes. (Section~\ref{sec:measurement_branch}; \href{https://github.com/Oxelra-AI/Qiushi-Engine-Babylm-Research/blob/main/research/materials/R18.md}{Materials: R18}.) \\[3pt]
R19\quad Self-sampled replaced-token detection & \textbf{Conditional result}. Add a detection objective and decompose its gradient effects. Short-run aggregate scores showed both gains and costs; overall effectiveness in late training remains unconfirmed. (Section~\ref{sec:measurement_branch}; \href{https://github.com/Oxelra-AI/Qiushi-Engine-Babylm-Research/blob/main/research/materials/R19.md}{Materials: R19}.) \\[3pt]
R20\quad Target weighting and supervision coverage & \textbf{Conditional result}. Compare per-word averaging, coverage floors, and source-absent content weighting. Local fit can improve at the cost of broader performance; replacement under an equal target budget still needs separate testing. (Section~\ref{sec:principles}; \href{https://github.com/Oxelra-AI/Qiushi-Engine-Babylm-Research/blob/main/research/materials/R20.md}{Materials: R20}.) \\[3pt]
R21\quad Relational pivots and counterfactual cues & \textbf{Conditional result}. Match supervision quantity and distinguish relational selection from direct retrieval. If retrieval still solves the task, high scores cannot establish a new computation for state selection. (Section~\ref{sec:principles}; \href{https://github.com/Oxelra-AI/Qiushi-Engine-Babylm-Research/blob/main/research/materials/R21.md}{Materials: R21}.) \\[3pt]
R22\quad Gradient exchange and projection & \textbf{Exploratory construction}. Use centered four-condition comparisons, directional projection, and negative examples. Local gradient differences have been observed, but long-term training gains have not been established. (Section~\ref{sec:measurement_branch}; \href{https://github.com/Oxelra-AI/Qiushi-Engine-Babylm-Research/blob/main/research/materials/R22.md}{Materials: R22}.) \\[3pt]
R23\quad Coherent-context margin loss & \textbf{Conditional result}. Compare against negatives with disrupted context and examine late-stage continuation. The expected behavioral separation was not obtained; the zero-coefficient comparison was not completed, so the result cannot be fully attributed to a single factor. (Section~\ref{sec:measurement_branch}; \href{https://github.com/Oxelra-AI/Qiushi-Engine-Babylm-Research/blob/main/research/materials/R23.md}{Materials: R23}.) \\[3pt]
\end{longtable}
\Needspace{8\baselineskip}
\subsection*{Curricula and learning dynamics}
\begin{longtable}{@{}P{39mm}P{111mm}@{}}
\toprule Topic & Intervention, finding, and status\\\midrule\endfirsthead
\toprule Topic (continued) & Intervention, finding, and status\\\midrule\endhead
\bottomrule\endfoot
R24\quad Adaptive masking granularity & \textbf{Experimental result}. Compare data-objective combinations along training trajectories. Early rankings can reverse; the learning stage must be considered when judging a training scheme. (Section~\ref{sec:frontier}; \href{https://github.com/Oxelra-AI/Qiushi-Engine-Babylm-Research/blob/main/research/materials/R24.md}{Materials: R24}.) \\[3pt]
R25\quad Data order and developmental curricula & \textbf{Experimental result}. Use random and developmental orderings of the same content and correct truncation. An effective combined scheme does not establish a universal advantage for curriculum ordering. (Section~\ref{sec:frontier}; \href{https://github.com/Oxelra-AI/Qiushi-Engine-Babylm-Research/blob/main/research/materials/R25.md}{Materials: R25}.) \\[3pt]
R26\quad Context-length caps & \textbf{Conditional result}. Hold paired-example exposure fixed while comparing length caps and evaluation lengths. The observed Overall change is affected by the AoA reference; the seven task families did not improve simultaneously. (\href{https://github.com/Oxelra-AI/Qiushi-Engine-Babylm-Research/blob/main/research/materials/R26.md}{Materials: R26}.) \\[3pt]
R27\quad Optimizers and mature model states & \textbf{Experimental result}. Compare AdamW, LAMB, optimizer switches at maturity, and trust ratios. Local numerical changes during optimization have not yet been shown to predict late-training model outcomes. (Section~\ref{sec:measurement_branch}; \href{https://github.com/Oxelra-AI/Qiushi-Engine-Babylm-Research/blob/main/research/materials/R27.md}{Materials: R27}.) \\[3pt]
R28\quad Checkpoint averaging and restarts & \textbf{Experimental result}. Compare averaging adjacent checkpoints, restarts, and another seed. Averaging did not reliably improve on individual high-scoring models; model selection and methodological attribution must be separated. (Section~\ref{sec:frontier}; \href{https://github.com/Oxelra-AI/Qiushi-Engine-Babylm-Research/blob/main/research/materials/R28.md}{Materials: R28}.) \\[3pt]
R29\quad Relation dose and learning maturity & \textbf{Experimental result}. Compare concentrated, sparse, and role-changing binding experience. Constructed tasks can be learned, but natural-text training with a weak base and sparse exposure did not automatically reproduce the capability; additional budget is listed separately. (Section~\ref{sec:principles}; \href{https://github.com/Oxelra-AI/Qiushi-Engine-Babylm-Research/blob/main/research/materials/R29.md}{Materials: R29}.) \\[3pt]
R30\quad Continued relation-training candidates & \textbf{Not run}. Retain proposals for continued relation losses, switching between local and full context, and remention masking. These remain routes to test, not achieved results. (Section~\ref{sec:principles}; \href{https://github.com/Oxelra-AI/Qiushi-Engine-Babylm-Research/blob/main/research/materials/R30.md}{Materials: R30}.) \\[3pt]
\end{longtable}
\Needspace{8\baselineskip}
\subsection*{Residual structures and incremental learning}
\begin{longtable}{@{}P{39mm}P{111mm}@{}}
\toprule Topic & Intervention, finding, and status\\\midrule\endfirsthead
\toprule Topic (continued) & Intervention, finding, and status\\\midrule\endhead
\bottomrule\endfoot
R31\quad Zero-initialized bottleneck branch & \textbf{Model method}. Check shared parameters, disabled-branch controls, and training trajectories. The branch preserves the base function at initialization and when disabled; joint training still updates the base. (Section~\ref{sec:frontier}; \href{https://github.com/Oxelra-AI/Qiushi-Engine-Babylm-Research/blob/main/research/materials/R31.md}{Materials: R31}.) \\[3pt]
R32\quad Main and auxiliary path separation & \textbf{Experimental result}. Move from broad to sparse auxiliary learning, including comparisons with the auxiliary branch disabled on the main path. Early gains did not carry over directly to mature models, motivating ordinary coherent replay. (Section~\ref{sec:frontier}; \href{https://github.com/Oxelra-AI/Qiushi-Engine-Babylm-Research/blob/main/research/materials/R32.md}{Materials: R32}.) \\[3pt]
R33\quad Dedicated increments on a frozen base & \textbf{Model method}. Train only newly added parameters, using ordinary prediction and KL; compare coherent and disrupted inputs. This produced the first-generation model; disabling the branch restores the base, but retention with it enabled still requires evaluation. (Section~\ref{sec:frontier}; \href{https://github.com/Oxelra-AI/Qiushi-Engine-Babylm-Research/blob/main/research/materials/R33.md}{Materials: R33}.) \\[3pt]
R34\quad Model lineage and scale selection & \textbf{Method or analysis tool}. Distinguish continuous-packing branches, the main model, incremental branches, and the inference scale used for release. Actual weight inheritance does not mean that every research route entered the final model in sequence. (Section~\ref{sec:lineage}; \href{https://github.com/Oxelra-AI/Qiushi-Engine-Babylm-Research/blob/main/research/materials/R34.md}{Materials: R34}.) \\[3pt]
\end{longtable}
\Needspace{8\baselineskip}
\subsection*{Entity memory and addressing}
\begin{longtable}{@{}P{39mm}P{111mm}@{}}
\toprule Topic & Intervention, finding, and status\\\midrule\endfirsthead
\toprule Topic (continued) & Intervention, finding, and status\\\midrule\endhead
\bottomrule\endfoot
R35\quad Writing and reading with supplied addresses & \textbf{Experimental result}. Compare consistent address permutations, wrong addresses, overwrites, and slot swaps. Stored signals have a functional role; results using gold-standard addresses do not establish natural-language addressing. (Section~\ref{sec:memory_branch}; \href{https://github.com/Oxelra-AI/Qiushi-Engine-Babylm-Research/blob/main/research/materials/R35.md}{Materials: R35}.) \\[3pt]
R36\quad Answer shortcuts in a tiny binding task & \textbf{Withdrawn}. Subsequent checks found that answers did not change with the queried entity. The original high scores no longer support compositional binding; correcting the corpus is not equivalent to verification by retraining. (Section~\ref{sec:memory_branch}; \href{https://github.com/Oxelra-AI/Qiushi-Engine-Babylm-Research/blob/main/research/materials/R36.md}{Materials: R36}.) \\[3pt]
R37\quad Routing from natural text to addresses & \textbf{Experimental result}. Compare raw-text, lexical, recurrent, and discrete routing with correct supplied addresses. High aggregate token accuracy concealed zero recall on critical queries, identifying an addressing bottleneck. (Section~\ref{sec:memory_branch}; \href{https://github.com/Oxelra-AI/Qiushi-Engine-Babylm-Research/blob/main/research/materials/R37.md}{Materials: R37}.) \\[3pt]
R38\quad Pointer and copy initialization & \textbf{Exploratory construction}. Test initialization for small pointer models, embeddings, and copying behavior. Being able to copy does not mean being able to select the correct entity state. (Section~\ref{sec:memory_branch}; \href{https://github.com/Oxelra-AI/Qiushi-Engine-Babylm-Research/blob/main/research/materials/R38.md}{Materials: R38}.) \\[3pt]
\end{longtable}
\Needspace{8\baselineskip}
\subsection*{Selection and readout}
\begin{longtable}{@{}P{39mm}P{111mm}@{}}
\toprule Topic & Intervention, finding, and status\\\midrule\endfirsthead
\toprule Topic (continued) & Intervention, finding, and status\\\midrule\endhead
\bottomrule\endfoot
R39\quad Format preadaptation and label alignment & \textbf{Experimental result}. Compare preparatory training with aligned, shuffled, and unrelated labels. Several forms of preparation can help subsequent learning, so the benefit cannot be attributed solely to one particular correspondence signal. (Section~\ref{sec:memory_branch}; \href{https://github.com/Oxelra-AI/Qiushi-Engine-Babylm-Research/blob/main/research/materials/R39.md}{Materials: R39}.) \\[3pt]
R40\quad Frozen readers and within-group selection & \textbf{Experimental result}. Train selection using a four-candidate within-group softmax, measuring lexical, role, and temporal semantics separately. Lexical matching and role swaps were nearly fully successful, but selecting prior versus updated states remained weak. (Section~\ref{sec:memory_branch}; \href{https://github.com/Oxelra-AI/Qiushi-Engine-Babylm-Research/blob/main/research/materials/R40.md}{Materials: R40}.) \\[3pt]
\end{longtable}
\Needspace{8\baselineskip}
\subsection*{Relational structure and identity}
\begin{longtable}{@{}P{39mm}P{111mm}@{}}
\toprule Topic & Intervention, finding, and status\\\midrule\endfirsthead
\toprule Topic (continued) & Intervention, finding, and status\\\midrule\endhead
\bottomrule\endfoot
R41\quad Identifiability of relational rules & \textbf{Experimental result}. Vary the initial owner and construct balanced counterexamples. High training scores may reflect an anti-copy shortcut; counterfactuals help identify the intended rule. (Section~\ref{sec:anchors}; \href{https://github.com/Oxelra-AI/Qiushi-Engine-Babylm-Research/blob/main/research/materials/R41.md}{Materials: R41}.) \\[3pt]
R42\quad Binary relation graphs and sparse anchors & \textbf{Experimental result}. Compare shared and separate representations, reverse anchors, and measure whether relative relations are preserved, using multiple training seeds. Absolute orientation can propagate along shared paths while relative relations are preserved; the conclusion is limited to the given hypothesis class. (Section~\ref{sec:anchors}; \href{https://github.com/Oxelra-AI/Qiushi-Engine-Babylm-Research/blob/main/research/materials/R42.md}{Materials: R42}.) \\[3pt]
R43\quad Character identity and access to relational structure & \textbf{Experimental result}. Learn matching from letter-equality supervision, then connect it to shared relational structure. Coverage of the base alphabet supports new combinations; candidates and hard assignments are supplied, and edge deletion also changes the training amount. (Section~\ref{sec:anchors}; \href{https://github.com/Oxelra-AI/Qiushi-Engine-Babylm-Research/blob/main/research/materials/R43.md}{Materials: R43}.) \\[3pt]
R44\quad Leakage-free graph and program construction & \textbf{Exploratory construction}. Generate neutral, reversed, and program-structured tasks, checking categories and priors. Candidate yield and type balance remain limited; prototypes do not establish general reasoning ability. (Section~\ref{sec:anchors}; \href{https://github.com/Oxelra-AI/Qiushi-Engine-Babylm-Research/blob/main/research/materials/R44.md}{Materials: R44}.) \\[3pt]
\end{longtable}
\Needspace{8\baselineskip}
\subsection*{Conditional value of experience}
\begin{longtable}{@{}P{39mm}P{111mm}@{}}
\toprule Topic & Intervention, finding, and status\\\midrule\endfirsthead
\toprule Topic (continued) & Intervention, finding, and status\\\midrule\endhead
\bottomrule\endfoot
R45\quad Structural density and representation recovery & \textbf{Conditional result}. Compare structural densities, residualized conditions, and random controls. Recovery of local structural readouts did not guarantee broad gains and must not be conflated with capacity. (Section~\ref{sec:experience_branch}; \href{https://github.com/Oxelra-AI/Qiushi-Engine-Babylm-Research/blob/main/research/materials/R45.md}{Materials: R45}.) \\[3pt]
R46\quad Tail training on shared-source clusters & \textbf{Conditional result}. Compare true, shuffled, repeated, and ordinary experience. Outperforming mismatched controls does not mean outperforming all references; model measurements are incomplete. (Section~\ref{sec:experience_branch}; \href{https://github.com/Oxelra-AI/Qiushi-Engine-Babylm-Research/blob/main/research/materials/R46.md}{Materials: R46}.) \\[3pt]
R47\quad Views, breadth, and repetition under a fixed budget & \textbf{Experimental result}. Replace training experience under a common budget and retain multiple late-stage models. Late-stage gains depend on experience type and architecture; adding distinct sentences is not equivalent to adding documents. (Section~\ref{sec:experience_branch}; \href{https://github.com/Oxelra-AI/Qiushi-Engine-Babylm-Research/blob/main/research/materials/R47.md}{Materials: R47}.) \\[3pt]
R48\quad Register substitution and data-value prediction & \textbf{Experimental result}. Hold added content fixed, vary the displaced register, and compare word distributions and task scores. Opportunity costs differ with the material displaced; results from different measurement settings cannot directly validate the same prediction. (Section~\ref{sec:experience_branch}; \href{https://github.com/Oxelra-AI/Qiushi-Engine-Babylm-Research/blob/main/research/materials/R48.md}{Materials: R48}.) \\[3pt]
R49\quad Selecting explicit relational text & \textbf{Exploratory construction}. Compare explicit relational text with random quality controls and filter natural state text. Qualified candidates are limited; short-run results cannot be treated as a mature large-scale training stream. (Section~\ref{sec:experience_branch}; \href{https://github.com/Oxelra-AI/Qiushi-Engine-Babylm-Research/blob/main/research/materials/R49.md}{Materials: R49}.) \\[3pt]
\end{longtable}
\Needspace{8\baselineskip}
\subsection*{Supervision targets and sources}
\begin{longtable}{@{}P{39mm}P{111mm}@{}}
\toprule Topic & Intervention, finding, and status\\\midrule\endfirsthead
\toprule Topic (continued) & Intervention, finding, and status\\\midrule\endhead
\bottomrule\endfoot
R50\quad Deletion of source-absent and copied targets & \textbf{Experimental result}. Keep inputs fixed and nearly match the number of deleted targets, using source-pair-disjoint and document-disjoint splits. Local source-pair results hold, but intervals under stricter document separation cross zero; rankings on the full task set differ. (Section~\ref{sec:principles}; \href{https://github.com/Oxelra-AI/Qiushi-Engine-Babylm-Research/blob/main/research/materials/R50.md}{Materials: R50}.) \\[3pt]
R51\quad Correct, wrong, and missing sources & \textbf{Experimental result}. Hold target tokens fixed and vary source conditions. Source effects can exist without added supervision and cannot directly be treated as the cause of final-model improvement. (Section~\ref{sec:relation}; \href{https://github.com/Oxelra-AI/Qiushi-Engine-Babylm-Research/blob/main/research/materials/R51.md}{Materials: R51}.) \\[3pt]
R52\quad Coverage, order, and local context & \textbf{Method or analysis tool}. Distinguish source presence, prefix placement, gaps, and relative text positions within windows. Comparisons with incompletely verified model-loading identity illustrate context geometry only; they do not confirm specific mechanism effects. (Section~\ref{sec:relation}; \href{https://github.com/Oxelra-AI/Qiushi-Engine-Babylm-Research/blob/main/research/materials/R52.md}{Materials: R52}.) \\[3pt]
\end{longtable}
\Needspace{8\baselineskip}
\subsection*{Functional mechanisms and learning signals}
\begin{longtable}{@{}P{39mm}P{111mm}@{}}
\toprule Topic & Intervention, finding, and status\\\midrule\endfirsthead
\toprule Topic (continued) & Intervention, finding, and status\\\midrule\endhead
\bottomrule\endfoot
R53\quad Ordered composition and dynamic binding & \textbf{Experimental result}. Compare dynamic and static relations, local and final supervision, and examine readouts. Information decodable from representations is not necessarily a computation the model actually performs. (Section~\ref{sec:reach}; \href{https://github.com/Oxelra-AI/Qiushi-Engine-Babylm-Research/blob/main/research/materials/R53.md}{Materials: R53}.) \\[3pt]
R54\quad Contextual hidden states and interventions & \textbf{Experimental result}. Compare consistent prefixes, mismatches, entity substitutions, and activation replacements. Topic information can be present even when state-dependent execution fails; behavioral and functional tests are both needed. (Section~\ref{sec:reach}; \href{https://github.com/Oxelra-AI/Qiushi-Engine-Babylm-Research/blob/main/research/materials/R54.md}{Materials: R54}.) \\[3pt]
R55\quad State updates in microworlds & \textbf{Method or analysis tool}. Control states, actions, contradictions, answer cues, and overwrites. These provide diagnostic tools, not automatic evidence that the same selector has formed in natural text. (Section~\ref{sec:reach}; \href{https://github.com/Oxelra-AI/Qiushi-Engine-Babylm-Research/blob/main/research/materials/R55.md}{Materials: R55}.) \\[3pt]
R56\quad Gradient conflict and learning-signal estimation & \textbf{Conditional result}. Use comparisons with identical materials, across pairs, and with held-out objects. Local correlations have not become stable prospective predictors of learning gains. (Section~\ref{sec:measurement_branch}; \href{https://github.com/Oxelra-AI/Qiushi-Engine-Babylm-Research/blob/main/research/materials/R56.md}{Materials: R56}.) \\[3pt]
R57\quad Contrastive prediction and auxiliary readouts & \textbf{Conditional result}. Compare same-source, cross-source, and hard-negative conditions. Fitting or saturating an auxiliary objective is insufficient evidence that new capability has formed. (Section~\ref{sec:measurement_branch}; \href{https://github.com/Oxelra-AI/Qiushi-Engine-Babylm-Research/blob/main/research/materials/R57.md}{Materials: R57}.) \\[3pt]
R58\quad Layer-wise readouts and position editing & \textbf{Conditional result}. Use bilinear and ridge-regression readouts, rotated-null controls, and local edits. Local fit does not guarantee held-out transfer; limited editing examples do not support broad mechanistic conclusions. (Section~\ref{sec:reach}; \href{https://github.com/Oxelra-AI/Qiushi-Engine-Babylm-Research/blob/main/research/materials/R58.md}{Materials: R58}.) \\[3pt]
R59\quad Score changes under pair swaps & \textbf{Method or analysis tool}. Check matching conditions including categories, shuffling, and length swaps. Algebraic cancellation in scores does not mean that the model actually used the correspondence. (Section~\ref{sec:measurement_branch}; \href{https://github.com/Oxelra-AI/Qiushi-Engine-Babylm-Research/blob/main/research/materials/R59.md}{Materials: R59}.) \\[3pt]
\end{longtable}
\Needspace{8\baselineskip}
\subsection*{Measurement, transfer, and attribution}
\begin{longtable}{@{}P{39mm}P{111mm}@{}}
\toprule Topic & Intervention, finding, and status\\\midrule\endfirsthead
\toprule Topic (continued) & Intervention, finding, and status\\\midrule\endhead
\bottomrule\endfoot
R60\quad Shared-parameter initialization & \textbf{Method or analysis tool}. Explicitly copy shared tensors and separately measure architectures' absolute capability and the relative effects of data. The same random seed does not guarantee identical initial values; after correction, evidence for a large interaction was substantially weakened. (Section~\ref{sec:measurement_branch}; \href{https://github.com/Oxelra-AI/Qiushi-Engine-Babylm-Research/blob/main/research/materials/R60.md}{Materials: R60}.) \\[3pt]
R61\quad Macro averages and item-level changes & \textbf{Method or analysis tool}. Compute gains, losses, net changes, and complementary items separately. Macro averages and net changes in the number of correct answers can have opposite signs; one aggregate score cannot fully describe capability changes. (Section~\ref{sec:measurement_branch}; \href{https://github.com/Oxelra-AI/Qiushi-Engine-Babylm-Research/blob/main/research/materials/R61.md}{Materials: R61}.) \\[3pt]
R62\quad Transfer conditions and representation geometry & \textbf{Conditional result}. Compare seed conditions, principal components, and random controls. Correlations and radial growth have not established capability conservation or a universal law of transfer. (Section~\ref{sec:measurement_branch}; \href{https://github.com/Oxelra-AI/Qiushi-Engine-Babylm-Research/blob/main/research/materials/R62.md}{Materials: R62}.) \\[3pt]
R63\quad Complete evaluation and model interfaces & \textbf{Method or analysis tool}. Verify the loading implementation, AoA trajectories, and commonsense-task and Reading scoring. Missing measurements, zero scores, and publicly displayed values are treated separately; complete model identity determines comparability. (Appendix~\ref{app:evaluation}; \href{https://github.com/Oxelra-AI/Qiushi-Engine-Babylm-Research/blob/main/research/materials/R63.md}{Materials: R63}.) \\[3pt]
\end{longtable}
\Needspace{8\baselineskip}
\subsection*{Systems and computation}
\begin{longtable}{@{}P{39mm}P{111mm}@{}}
\toprule Topic & Intervention, finding, and status\\\midrule\endfirsthead
\toprule Topic (continued) & Intervention, finding, and status\\\midrule\endhead
\bottomrule\endfoot
R64\quad Word boundaries, packing, and visible budgets & \textbf{Method or analysis tool}. Check pair integrity, truncation, and actual presentation in continuous windows. Accounted word counts must correspond to visible content; prototype and final tokenizer implementations must not be mixed. (Section~\ref{sec:measurement_branch}; \href{https://github.com/Oxelra-AI/Qiushi-Engine-Babylm-Research/blob/main/research/materials/R64.md}{Materials: R64}.) \\[3pt]
R65\quad Microbatches and gradient equivalence & \textbf{Method or analysis tool}. Weight by target count and compare gradients before and after activation recomputation under fixed randomness. These checks validate limited implementation conditions, not exact identity of complete stochastic training trajectories. (Section~\ref{sec:measurement_branch}; \href{https://github.com/Oxelra-AI/Qiushi-Engine-Babylm-Research/blob/main/research/materials/R65.md}{Materials: R65}.) \\[3pt]
R66\quad Model reconstruction and cost breakdown & \textbf{Method or analysis tool}. Fix the identities of weights, configuration, and code, separating training from search costs. Same-seed reconstruction, replication across seeds, and compute savings are distinct conclusions. (Appendix~\ref{app:repro}; \href{https://github.com/Oxelra-AI/Qiushi-Engine-Babylm-Research/blob/main/research/materials/R66.md}{Materials: R66}.) \\[3pt]
\end{longtable}
\Needspace{8\baselineskip}
\subsection*{Core research: extensions and empirical tests}
\begin{longtable}{@{}P{39mm}P{111mm}@{}}
\toprule Topic & Intervention, finding, and status\\\midrule\endfirsthead
\toprule Topic (continued) & Intervention, finding, and status\\\midrule\endhead
\bottomrule\endfoot
R67\quad Selectivity in relation learning & \textbf{Experimental result}. Compare source advantages from verbatim repetition and aligned reformulation across three seeds. The effects on compact reformulation targets differ in direction and do not imply that repetition is universally harmful. (Section~\ref{sec:relation}; \href{https://github.com/Oxelra-AI/Qiushi-Engine-Babylm-Research/blob/main/research/materials/R67.md}{Materials: R67}.) \\[3pt]
R68\quad Prediction windows with matched materials & \textbf{Experimental result}. Split repetition pairs and reformulation pairs into separate windows in their respective conditions. The large source effects were substantially attenuated; content is matched between the same-window and split-window versions of each pairing condition. (Section~\ref{sec:relation}; \href{https://github.com/Oxelra-AI/Qiushi-Engine-Babylm-Research/blob/main/research/materials/R68.md}{Materials: R68}.) \\[3pt]
R69\quad Transfer conditions for natural reformulations & \textbf{Experimental result}. Group natural-restatement targets by whether their exact tokenizer ID appears in the source. Aligned restatement improves source use for recurring target tokens; nonrecurring targets show no equally stable improvement over the reference. This distinction is lexical, not a definition of semantic novelty. (Section~\ref{sec:relation}; \href{https://github.com/Oxelra-AI/Qiushi-Engine-Babylm-Research/blob/main/research/materials/R69.md}{Materials: R69}.) \\[3pt]
R70\quad Familiar performance and functional access with unseen symbols & \textbf{Experimental result}. Compare ordinary continuation, statically weighted supervision, and interleaved supervision. Recovery of familiar behavior does not guarantee restored functional reach for unseen symbols. (Section~\ref{sec:reach}; \href{https://github.com/Oxelra-AI/Qiushi-Engine-Babylm-Research/blob/main/research/materials/R70.md}{Materials: R70}.) \\[3pt]
R71\quad Erasing and relocating functional signals & \textbf{Experimental result}. Center, zero, and rotate attribute-position signals, and test refitted directions. The results support the functional role of a specific signal; unsuccessful recovery using linear directions does not rule out all distributed recoding. (Section~\ref{sec:reach}; \href{https://github.com/Oxelra-AI/Qiushi-Engine-Babylm-Research/blob/main/research/materials/R71.md}{Materials: R71}.) \\[3pt]
R72\quad Dense inputs and sparse supervision & \textbf{Model method}. Reuse existing text while controlling masking candidates and focused targets separately. At equal exposure, complete evaluations for both seeds outperformed ordinary continuation; the comparison does not isolate masking as a single factor. (Section~\ref{sec:guided}; \href{https://github.com/Oxelra-AI/Qiushi-Engine-Babylm-Research/blob/main/research/materials/R72.md}{Materials: R72}.) \\[3pt]
R73\quad Preservation inputs and effective strength & \textbf{Experimental result}. Measure preservation gradients at fixed shared target positions and compare ordinary and dense inputs. With the same coefficient, gradient magnitudes still differ by a factor of approximately 12; the full comparison also changes the set of target positions. (Section~\ref{sec:guided}; \href{https://github.com/Oxelra-AI/Qiushi-Engine-Babylm-Research/blob/main/research/materials/R73.md}{Materials: R73}.) \\[3pt]
R74\quad Principle-guided complete model & \textbf{Model method}. Recheck ordinary continuation, relation learning, and the complete preservation scheme from the same parent model. The complete scheme improved the nine-task Overall score for both continuation seeds; additional exposure and compute are reported separately. (Section~\ref{sec:complete_results}; \href{https://github.com/Oxelra-AI/Qiushi-Engine-Babylm-Research/blob/main/research/materials/R74.md}{Materials: R74}.) \\[3pt]
\end{longtable}